\documentclass[letterpaper]{article} 
\PassOptionsToPackage{table}{xcolor}
\usepackage[preprint]{aaai2027}
\usepackage[hyphens]{url} 
\usepackage{graphicx} 
\usepackage{natbib} 
\usepackage{caption} 
\usepackage{amsmath}
\usepackage{amssymb}
\usepackage{latexsym}
\usepackage{booktabs}
\usepackage[most]{tcolorbox}
\usepackage{algorithm}
\usepackage{algorithmic}

\title{R3S: Refining and Recovering Reinforcement Signals for Multilingual Understanding and Reasoning}

\author{
    Junxiao Liu\textsuperscript{\rm 1},
    Zhijun Wang\textsuperscript{\rm 1},
    Yixiao Li\textsuperscript{\rm 1},
    Zhejian Lai\textsuperscript{\rm 1},
    Liqian Huang\textsuperscript{\rm 2},
    Xin Huang\textsuperscript{\rm 3}\corresponding,\\
    Xue Han\textsuperscript{\rm 3},
    Junlan Feng\textsuperscript{\rm 3},
    Shujian Huang\textsuperscript{\rm 1}\corresponding
}
\affiliations{
    \textsuperscript{\rm 1}National Key Laboratory for Novel Software Technology, Nanjing University\\
    \textsuperscript{\rm 2}University of Tübingen\\
    \textsuperscript{\rm 3}China Mobile Communications Company Limited Research Institute\\
    \{junxiaoliu, wangzj, liyixiao, laizj\}@smail.nju.edu.cn,
    liqian.huang@student.uni-tuebingen.de,\\
    \{huangxin, hanxuejt\}@cmjt.chinamobile.com,
    fengjunlan@chinamobile.com,
    huangsj@nju.edu.cn
}

\newcommand{\method}{R3S}
\begin{document}
\maketitle

\begin{abstract}

Large reasoning models often default to English reasoning when processing non-English questions, yet their performance drops substantially when reasoning in the question language. Even with the same reasoning language, semantically equivalent English and non-English questions still exhibit a clear performance gap. Together, these phenomena reveal two distinct bottlenecks: target-language question understanding and target-language reasoning. Existing methods typically optimize only one of these capabilities. However, simply combining them may not be sufficient to optimize both effectively, as answer correctness alone cannot distinguish failures in question understanding from those in reasoning. We propose \textbf{\method{}}, a reinforcement learning framework that disentangles the optimization of the two capabilities. \method{} refines translation rewards derived from downstream reasoning accuracy through English-solvability filtering and recovers target-language RLVR signals using self-generated English hints. Together, these designs require neither external model feedback nor external multilingual training data. Experiments across three backbone models and five languages show that \method{} improves language-consistent accuracy over the target-language RLVR baseline on MMATH by an average of 10.3 percentage points, while maintaining near-perfect language consistency. Consistent gains on MMLU-ProX further demonstrate its generalization beyond math problems.
\end{abstract}

\section{Introduction}

Large reasoning models (LRMs), typically trained via reinforcement learning with verifiable rewards (RLVR)~\cite{deepseekr1}, have achieved strong performance on complex reasoning tasks under the ``think-then-answer'' paradigm~\cite{qwen3,o3}. However, these capabilities vary substantially across languages. When presented with non-English questions, models often default to English reasoning, creating a mismatch between the input and reasoning languages. Requiring them to reason in the question language typically causes a pronounced performance drop accompanied by degenerate repetition, indicating limited \textbf{target-language reasoning} ability~\cite{Qi_2025,polymath}. Furthermore, even when the reasoning language is held fixed, models still exhibit a substantial performance gap between semantically equivalent English and non-English questions, indicating limitations in \textbf{target-language question understanding}~\cite{ko2025understandsolvetranslatebridging,kang2026multilingualreasoninggapsemerge}. Taken together, these patterns highlight two bottlenecks in multilingual reasoning: target-language question understanding and target-language reasoning.

Existing methods typically optimize one of these two capabilities. Translation-based methods, such as QAlign~\cite{zhu2024questiontranslationtrainingbetter} and TAPO~\cite{tapo}, establish semantic correspondence across languages to improve the understanding of non-English questions. However, their reasoning process remains primarily in English, so they do not directly improve the model's ability to reason in the question language. In contrast, M-Thinker~\cite{thinkNatively} and MAPO~\cite{mapo} align target-language reasoning trajectories with their English counterparts. These methods improve target-language reasoning and language consistency, but do not explicitly optimize target-language question understanding or address the limited exploration of correct reasoning trajectories in the target language. In practice, however, \textbf{the two capabilities are entangled}: the outcome of reasoning over a target-language question depends on both whether the question is understood and whether the model can reason in that language. Consequently, optimizing either capability in isolation can be affected by limitations in the other. 

To this end, we propose \textbf{\method{}} (\textbf{R}efining and \textbf{R}ecovering \textbf{R}einforcement \textbf{S}ignals), a reinforcement learning framework that optimizes target-language question understanding and target-language reasoning with capability-specific reinforcement signals. For target-language question understanding, \method{} translates English problems into target languages and uses downstream reasoning accuracy as the translation reward. To make the reward more accurate, \method{} restricts translation training to questions that the model can reliably solve in English. This selection reduces cases in which an incorrect response is caused by insufficient problem-solving ability rather than a translation error, producing a \textbf{refined} training signal for translation.

For target-language reasoning, \method{} performs target-language RLVR, but only on translated questions that the model is able to solve in English. This selection reduces the influence of question misunderstanding and insufficient problem-solving ability, \textbf{refining} the training signal to be more specific to target-language reasoning. When direct sampling produces no correct response in the target language, leaving no effective signal for learning correct reasoning, \method{} uses self-generated English reasoning hints to guide target-language exploration and \textbf{recover} effective signals.

Moreover, \method{} performs English RLVR on all source questions to improve the model's problem-solving ability. Since translation feedback comes from verifiable downstream outcomes and reasoning hints are self-generated, \method{} requires no external annotation or model feedback.

Experiments across five non-English languages on three models with varying levels of multilingual capability demonstrate the effectiveness of \method{}. On MMATH, \method{} outperforms the target-language RLVR baseline by an average of 10.3 percentage points while maintaining near-perfect language consistency. Similar gains on MMLU-ProX show the effectiveness of \method{} beyond math problems. Further analysis shows that English-solvability filtering substantially reduces noise in translation rewards, validating the reliability of the refined translation signal.

Our contributions are threefold.
(1) We propose \method{}, a reinforcement learning framework that improves target-language question understanding and target-language reasoning by refining translation rewards through English-solvability filtering and recovering effective target-language RLVR signals with self-generated English reasoning hints.
(2) Starting only from English problems, \method{} constructs multilingual training instances and reasoning hints from the model's own generations, without external model feedback.
(3) Across three backbone models, \method{} consistently outperforms SoftLC-RL on MMATH, while consistent improvements on MMLU-ProX demonstrate generalization beyond math problems.





\section{Related Work}

Although large reasoning models perform strongly on English reasoning tasks, their multilingual reasoning remains limited~\cite{Qi_2025,polymath,breakingLanguageBarrier}. Existing work has sought to improve multilingual reasoning from two main perspectives: multilingual question understanding and target-language reasoning. On the question-understanding side, prior studies~\cite{ko2025understandsolvetranslatebridging,kang2026multilingualreasoninggapsemerge} show that even when the reasoning language is fixed, model performance varies substantially across semantically equivalent questions expressed in different languages. Translation-based methods address this issue by establishing semantic correspondence across languages. QAlign~\cite{zhu2024questiontranslationtrainingbetter} trains models to translate non-English questions before reasoning in English, while TAPO~\cite{tapo} uses translation training to improve reasoning over multilingual questions. These methods improve the understanding of non-English questions, but their reasoning process remains primarily in English and therefore does not directly improve target-language reasoning.

On the target-language reasoning side, supervised fine-tuning with translated chain-of-thought data~\cite{breakingLanguageBarrier} teaches models to generate reasoning trajectories in the question language. M-Thinker~\cite{thinkNatively} and MAPO~\cite{mapo} transfer English reasoning ability by aligning target-language reasoning trajectories with their English counterparts, while related preference optimization and reinforcement learning methods improve language control and cross-lingual reasoning transfer~\cite{park2025crosslingualcollapselanguagecentricfoundation,hwang2025learngloballyspeaklocally}. These methods improve target-language reasoning and language consistency, but they do not explicitly optimize multilingual question understanding or address the limited exploration of correct reasoning trajectories in the target language.

In contrast, \method{} optimizes multilingual question understanding and target-language reasoning using distinct reinforcement signals. For question understanding, it uses English-solvability filtering to refine translation rewards derived from downstream reasoning accuracy. For target-language reasoning, it uses self-generated English reasoning hints to recover effective RLVR signals when all directly sampled target-language responses are incorrect. Both signals are generated by the model itself: translation feedback is derived from verifiable downstream outcomes, and reasoning hints are self-generated. Consequently, during its two reinforcement-learning stages, \method{} starts from English source problems alone and requires neither pre-constructed multilingual training data nor external model feedback.

\section{Methods}
\begin{figure*}[!t]
\centering
\includegraphics[width=\linewidth]{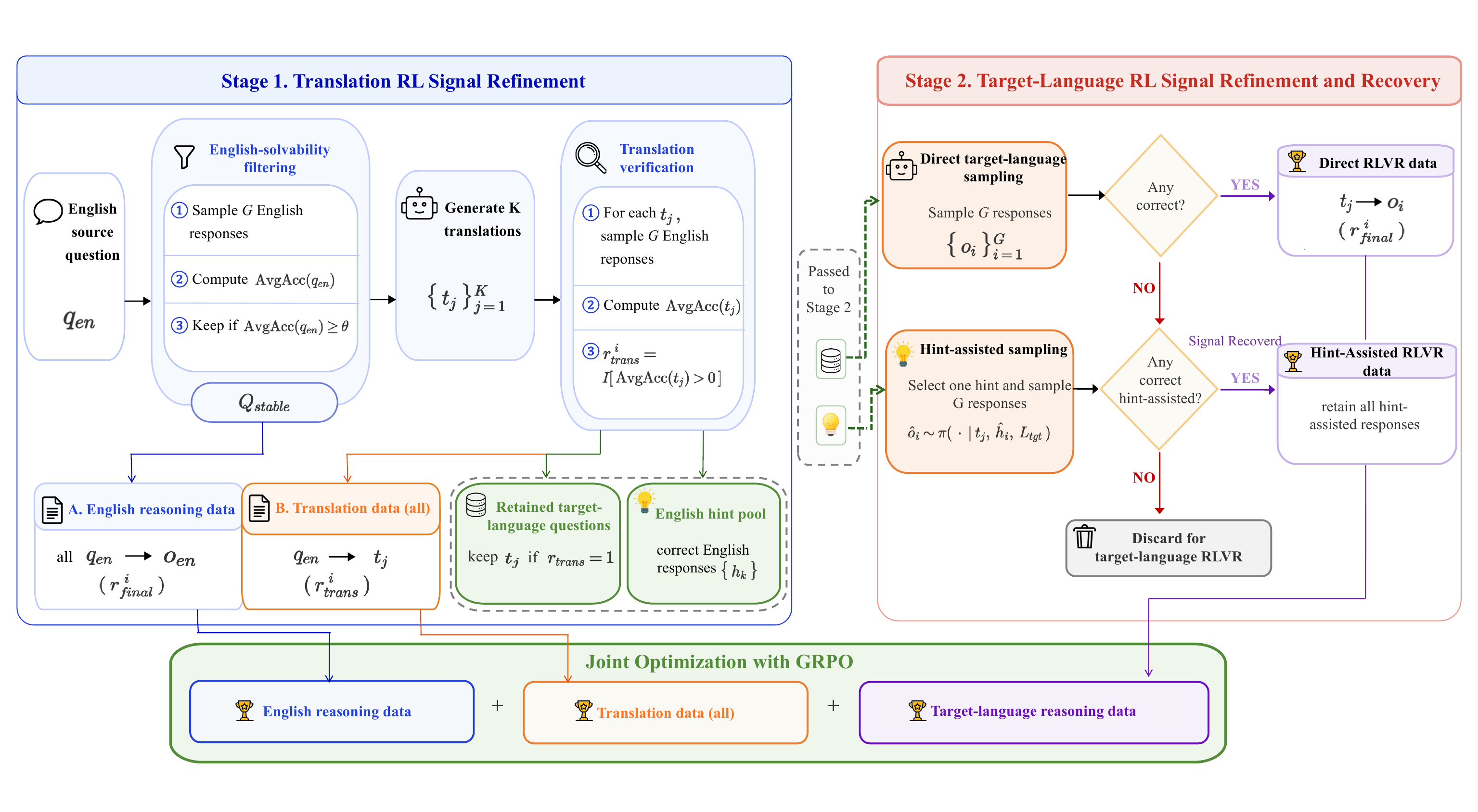}
\caption{
    Overview of \method.
    Stage 1 refines translation RL signals through English-solvability filtering. Stage 2 uses English-verified translations to refine target-language RLVR signals and self-generated English hints to recover signals when direct sampling fails. The resulting English reasoning, translation, and target-language reasoning data are used for GRPO optimization.
}
\label{fig:framework}
\end{figure*}

\begin{figure}[!ht]
    \centering
    \includegraphics[width=\linewidth]{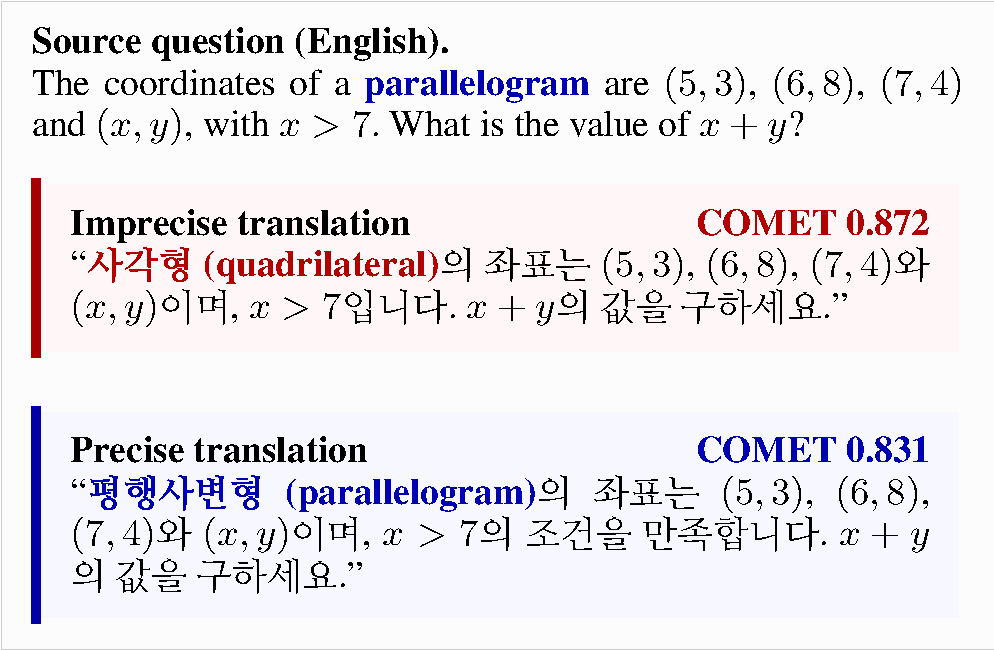}
    \caption{
        A task-critical semantic error missed by COMET.
        COMET assigns a higher score to an imprecise translation
        (``parallelogram''$\rightarrow$``quadrilateral,'' 0.872) than to
        the precise translation (0.831), despite the loss of a mathematical
        constraint.
    }
    \label{fig:case_parallelogram}
\end{figure}

We propose \method{}, a reinforcement learning framework that optimizes multilingual question understanding and target-language reasoning using distinct reinforcement signals. For multilingual question understanding, \method{} uses downstream reasoning accuracy to reward target-language translations and refines the translation reward through English-solvability filtering. For target-language reasoning, \method{} uses only translated problems that the model has already solved in English, reducing interference from question misunderstanding and insufficient problem-solving ability and thereby refining the target-language RLVR signal. When direct sampling yields no correct response, \method{} uses self-generated English reasoning hints to recover an effective signal. Figure~\ref{fig:framework} provides an overview of \method{}, while Algorithm~\ref{alg:trit} in Appendix~\ref{app:algorithm} details the complete training procedure.

\subsection{Translation Reinforcement Signal Refinement}

To improve multilingual question understanding, we use translation training to establish semantic correspondence between equivalent questions across languages. Given an English problem with a verifiable answer, the model generates $K$ candidate translations in the target language.

A key challenge is accurately evaluating the generated translations. Mathematical problems contain precise terminology and strict constraints, and even a subtle mistranslation can change the problem and its solution. As shown in Figure~\ref{fig:case_parallelogram}, an incorrect translation renders ``parallelogram'' as ``quadrilateral,'' removing a geometric constraint essential to solving the problem. However, COMET~\cite{comet} assigns the incorrect translation a score of $0.872$, higher than the $0.831$ assigned to the accurate translation. This example illustrates that general-purpose translation metrics can be insensitive to subtle changes in task-critical semantics.

Motivated by this observation, we use downstream reasoning accuracy to construct a binary translation reward. For each candidate translation $t_j$, the model generates $G$ English responses to the translated problem and computes their average accuracy. We define its binary translation reward as:

\begin{equation}
r_{\mathrm{trans}}^{j}
=
\mathbb{I}\!\left[\operatorname{AvgAcc}(t_j)>0\right].
\end{equation}

Thus, a translation receives a positive reward if it satisfies the language and format requirements and at least one English response is correct; otherwise, its reward is zero. This binary reward does not rank translations by surface-level similarity; instead, it indicates whether a translation preserves enough task-critical information for the problem to remain solvable.

However, when all $G$ English responses are incorrect, a zero translation reward is ambiguous: the translation may be inaccurate, or the model may be unable to solve the underlying problem. \method{} therefore refines the translation reward through English-solvability filtering. For each training batch of English source problems $\mathcal{Q}_{\mathrm{en}}$, the model first generates $G$ English responses to every problem and computes their average accuracy. We retain only source problems whose average accuracy is no lower than a threshold $\theta$:

\begin{equation}
\mathcal{Q}_{\mathrm{stable}}
=
\left\{
q_{\mathrm{en}}\in\mathcal{Q}_{\mathrm{en}}
\;\middle|\;
\operatorname{AvgAcc}(q_{\mathrm{en}})\geq\theta
\right\}.
\end{equation}

Only problems in $\mathcal{Q}_{\mathrm{stable}}$ are used for translation training. This restriction reduces zero rewards caused by insufficient problem-solving ability, thereby refining the translation reward. Our error analysis further shows that English-solvability filtering reduces the false-negative rate of translation rewards from $31.3\%$ to $5.2\%$ (Appendix~\ref{app:translation_reward_noise}).

To prevent translation training from being confined to currently solvable problems, \method{} performs English RLVR on all source problems. As the model's English reasoning improves, more problems can pass the solvability filter and enter translation training in subsequent batches.

All candidate translations and their rewards are used for translation training. Positively rewarded translations are retained as target-language problems for the next stage, and all correct English responses generated during translation verification are retained for constructing reasoning hints.

\subsection{Target-Language Reinforcement Signal Recovery}

To improve target-language reasoning, we use only the target-language problems retained in the preceding stage. Since the model has answered each problem correctly in English, failures on these problems are less likely to result from question misunderstanding or insufficient problem-solving ability. Restricting target-language RLVR to these verified problems therefore refines the reinforcement signal to be more specific to target-language reasoning.

For each problem, the model first samples $G$ responses directly in the target language without any hints. If at least one response is correct, all sampled responses and their rewards are retained for target-language RLVR. For some problems, however, all $G$ responses are incorrect even though the model has already solved the problem in English. In these cases, the model fails to explore a correct target-language reasoning trajectory. With no correct trajectory to reinforce, RLVR provides no effective signal for learning correct reasoning.

To recover an effective reinforcement signal for these problems, \method{} constructs a hint pool from the correct English responses generated when computing the translation rewards. We extract an approximately $200$-token prefix from each correct response, ending at the nearest sentence boundary, as an English reasoning hint. When all direct target-language responses are incorrect, we uniformly sample one hint $h$ from the available hint pool and generate a new group of $G$ responses conditioned on the same hint. Given the target-language problem $t_j$ and the selected hint $h$, the current policy samples each hint-assisted response as

\begin{equation}
\widetilde{o}_i
\sim
\pi\!\left(\cdot \mid t_j,h,L_{\mathrm{tgt}}\right),
\qquad i=1,\ldots,G,
\end{equation}

where $L_{\mathrm{tgt}}$ denotes the instruction to respond entirely in the target language, and $\widetilde{o}_i$ is the resulting complete response. If at least one hint-assisted response is correct, \method{} recovers an effective reinforcement signal for the problem, and all hint-assisted responses and their rewards are retained for RLVR; otherwise, they are discarded. For retained hint-assisted samples, the English hint remains part of the input during policy optimization.

Within each training batch, \method{} uses GRPO~\cite{grpo} to update the model with English reasoning, translation, and target-language reasoning data. Since translation feedback comes from verifiable downstream outcomes and reasoning hints are self-generated, \method{} requires no external model feedback.

For target-language reasoning, we use a compositional reward that accounts for answer correctness, language consistency, output format, and degenerate repetition. We assign a reward of 1 to correct responses satisfying all quality constraints, 0.1 to incorrect but otherwise valid responses, and 0 otherwise. The complete reward definition and implementation details are provided in Appendix~\ref{app:repetition}.

\section{Experiments}

\begin{table*}[t]
\centering
{\small
\setlength{\tabcolsep}{1mm}
\begin{tabular}{@{}l*{5}{cc}ccc c@{}}
\toprule
& \multicolumn{2}{c}{\textbf{FR}}
& \multicolumn{2}{c}{\textbf{PT}}
& \multicolumn{2}{c}{\textbf{JA}}
& \multicolumn{2}{c}{\textbf{KO}}
& \multicolumn{2}{c}{\textbf{TH}}
& \multicolumn{3}{c}{\textbf{Non-EN Avg.}}
& \textbf{EN} \\
\cmidrule(lr){2-3}\cmidrule(lr){4-5}\cmidrule(lr){6-7}
\cmidrule(lr){8-9}\cmidrule(lr){10-11}\cmidrule(lr){12-14}
\textbf{Method}
& \textbf{LC\&Acc} & \textbf{Acc.}
& \textbf{LC\&Acc} & \textbf{Acc.}
& \textbf{LC\&Acc} & \textbf{Acc.}
& \textbf{LC\&Acc} & \textbf{Acc.}
& \textbf{LC\&Acc} & \textbf{Acc.}
& \textbf{LC\&Acc} & \textbf{Acc.} & \textbf{LC}
& \textbf{LC\&Acc} \\
\midrule
\multicolumn{15}{@{}l}{\textbf{DeepSeek-Distill-Qwen-1.5B}} \\
Base
& 6.3 & 34.8 & 10.3 & 34.4 & 0.3 & 30.4 & 0.1 & 32.2 & 0.4 & 24.4 & 3.5 & 31.2 & 19.3 & 42.9 \\
SFT
& 22.7 & 22.7 & 24.2 & 24.5 & 11.5 & 11.5 & 10.1 & 10.3 & 9.6 & 9.6 & 15.6 & 15.7 & 97.9 & 38.5 \\
Naive RL
& 2.0 & \textbf{46.5} & 0.0 & \textbf{45.3} & 0.0 & \textbf{40.6} & 0.0 & \textbf{39.7} & 0.0 & \textbf{37.4} & 0.4 & \textbf{41.9} & 1.3 & 47.6 \\
SoftLC-RL
& 36.4 & 36.4 & 38.0 & 38.1 & 22.7 & 22.7 & 0.0 & 38.9 & 23.6 & 23.7 & 24.1 & 32.0 & 81.0 & 48.4 \\
M-Thinker
& 35.6 & 35.6 & 33.9 & 34.3 & 30.1 & 30.1 & 23.6 & 23.7 & 25.7 & 26.0 & 29.8 & 29.9 & 99.7 & 38.9 \\
Direct Comb.
& 38.4 & 39.0 & 35.1 & 35.1 & 29.4 & 29.4 & 20.4 & 20.4 & 27.5 & 27.8 & 30.1 & 30.3 & 99.2 & 49.7 \\
\method{}
& \textbf{45.7} & 45.9 & \textbf{43.0} & 43.0 & \textbf{34.0} & 34.0 & \textbf{26.2} & 26.2 & \textbf{31.5} & 31.5 & \textbf{36.1} & 36.2 & \textbf{99.9} & \textbf{52.0} \\
\midrule
\multicolumn{15}{@{}l}{\textbf{Qwen3-1.7B}} \\
Base
& 0.0 & 42.8 & 0.0 & 43.3 & 0.0 & 40.7 & 0.0 & 41.2 & 0.0 & 40.0 & 0.0 & 41.6 & 0.0 & 41.7 \\
SFT
& 35.0 & 37.2 & 36.4 & 36.5 & 25.6 & 25.6 & 24.6 & 25.0 & 25.5 & 25.6 & 29.4 & 30.0 & 98.8 & 34.4 \\
Naive RL
& 0.0 & 50.5 & 0.0 & 51.1 & 0.0 & \textbf{46.4} & 0.0 & \textbf{45.9} & 0.0 & \textbf{46.8} & 0.0 & \textbf{48.1} & 0.0 & 54.5 \\
SoftLC-RL
& 40.6 & 40.6 & 41.3 & 41.3 & 32.0 & 32.0 & 34.4 & 34.4 & 35.0 & 35.0 & 36.7 & 36.7 & 99.9 & 39.7 \\
M-Thinker
& 42.0 & 42.0 & 43.1 & 43.1 & 34.0 & 34.0 & 31.1 & 31.1 & 34.0 & 34.0 & 36.8 & 36.8 & 99.9 & 47.4 \\
Direct Comb.
& 46.4 & 46.5 & 51.4 & 51.4 & 41.5 & 41.5 & 37.8 & 37.8 & 43.1 & 43.1 & 44.1 & 44.1 & 99.9 & 55.8 \\
\method{}
& \textbf{53.7} & \textbf{53.7} & \textbf{53.3} & \textbf{53.4} & \textbf{45.8} & 45.8 & \textbf{39.9} & 39.9 & \textbf{44.2} & 44.2 & \textbf{47.4} & 47.4 & \textbf{100.0} & \textbf{58.1} \\
\midrule
\multicolumn{15}{@{}l}{\textbf{Qwen3-4B}} \\
Base
& 0.0 & 53.2 & 0.0 & 53.3 & 0.0 & 51.8 & 0.0 & 52.3 & 0.0 & 50.9 & 0.0 & 52.3 & 0.0 & 51.4 \\
SFT
& 37.5 & 38.0 & 38.3 & 38.3 & 25.6 & 25.6 & 25.2 & 25.2 & 19.9 & 19.9 & 29.3 & 29.4 & 99.4 & 46.7 \\
Naive RL
& 0.0 & 65.1 & 0.0 & 64.3 & 0.0 & 60.4 & 0.0 & \textbf{62.7} & 0.0 & 62.3 & 0.0 & 63.0 & 0.0 & 65.8 \\
SoftLC-RL
& 60.9 & 60.9 & 63.2 & 63.2 & 51.7 & 51.8 & 48.9 & 48.9 & 53.0 & 53.0 & 55.5 & 55.6 & 99.9 & 39.7 \\
M-Thinker
& 60.8 & 60.8 & 60.5 & 60.5 & 51.9 & 51.9 & 52.9 & 53.0 & 53.3 & 53.3 & 55.9 & 55.9 & 99.9 & 25.2 \\
Direct Comb.
& 61.0 & 61.0 & 65.6 & 65.6 & 58.3 & 58.3 & 57.1 & 57.1 & 59.5 & 59.5 & 60.3 & 60.3 & 100.0 & 67.4 \\
\method{}
& \textbf{66.9} & \textbf{66.9} & \textbf{66.2} & \textbf{66.2} & \textbf{63.2} & \textbf{63.2} & \textbf{58.5} & 58.5 & \textbf{63.6} & \textbf{63.6} & \textbf{63.7} & \textbf{63.7} & \textbf{100.0} & \textbf{68.8} \\
\bottomrule
\end{tabular}
}
\caption{Main results on MMATH (\%; avg@4). The best result within each backbone and column is shown in bold.}
\label{tab:main_result}
\end{table*}

\subsection{Experiment Setup}

\paragraph{Backbone Models.}
We evaluate \method{} on three backbones. The first is DeepSeek-Distill-Qwen-1.5B; the other two are Qwen3-1.7B and Qwen3-4B. Together, they span different model scales and levels of multilingual capability, allowing us to assess the robustness of \method{} across backbone settings.

\paragraph{Benchmarks.}
Following prior work on target-language reasoning~\cite{thinkNatively}, we evaluate five target languages: French, Portuguese, Japanese, Korean, and Thai. We use two complementary multilingual benchmarks, MMATH~\cite{mmath} and MMLU-ProX~\cite{mmluprox}. MMATH comprises four mathematical reasoning subsets of varying difficulty: MATH500, CNMO, AIME24, and AIME25. To assess generalization beyond the mathematical training domain, we additionally evaluate on five STEM domains from MMLU-ProX: biology, chemistry, physics, engineering, and mathematics.

\paragraph{Evaluation Metrics.}
For each problem, we sample four responses and report their average performance (avg@4). For each target language, we macro-average the results across the four MMATH subsets and the five MMLU-ProX STEM domains. Following prior work on multilingual reasoning~\cite{thinkNatively}, we report three metrics: \textbf{Language Consistency (LC)}, which measures whether the reasoning trace is in the question language; \textbf{Accuracy (Acc)}, which measures final-answer correctness; and \textbf{LC\&Acc}, which measures the percentage of responses that are both correct and language-consistent. We use LC\&Acc as our primary metric.

\paragraph{Baselines.}
We compare \method{} against five baselines:
\begin{itemize}
    \item \textbf{SFT:} Fine-tunes the model on target-language question--response pairs generated by Qwen3-32B.

    \item \textbf{Naive RL:} Optimizes only response correctness using the accuracy reward ($r_{\mathrm{acc}}$), without enforcing language consistency.

    \item \textbf{SoftLC-RL}~\cite{mistralai2025magistral}: Adds a soft language reward of 0.1 to Naive RL when the output is in the target language.

    \item \textbf{M-Thinker}~\cite{thinkNatively}: Uses language-consistency and cross-lingual thinking-alignment rewards from an external model to align multilingual reasoning traces with English.

    \item \textbf{Direct Combination:} Jointly trains translation and target-language RLVR under the same data and optimization budget as \method{}, but uses neither English-solvability filtering nor hint-based signal recovery.
\end{itemize}

For each backbone, all RL-based methods start from a common checkpoint to ensure a fair comparison. All questions are drawn from DAPO-MATH-17K~\cite{dapo-math}. For baselines requiring multilingual training data, we translate the source questions into the target languages with DeepSeek-V3.2-Exp and apply a multi-pass verification procedure. Details on data construction and implementation are provided in Appendix~\ref{app:training_details}.

\subsection{Experiment Results}

\begin{table*}[t]
\centering
{\small
\setlength{\tabcolsep}{1mm}
\begin{tabular}{@{}l*{5}{cc}ccc c@{}}
\toprule
& \multicolumn{2}{c}{\textbf{FR}}
& \multicolumn{2}{c}{\textbf{PT}}
& \multicolumn{2}{c}{\textbf{JA}}
& \multicolumn{2}{c}{\textbf{KO}}
& \multicolumn{2}{c}{\textbf{TH}}
& \multicolumn{3}{c}{\textbf{Non-EN Avg.}}
& \textbf{EN} \\
\cmidrule(lr){2-3}\cmidrule(lr){4-5}\cmidrule(lr){6-7}
\cmidrule(lr){8-9}\cmidrule(lr){10-11}\cmidrule(lr){12-14}
\textbf{Method}
& \textbf{LC\&Acc} & \textbf{Acc.}
& \textbf{LC\&Acc} & \textbf{Acc.}
& \textbf{LC\&Acc} & \textbf{Acc.}
& \textbf{LC\&Acc} & \textbf{Acc.}
& \textbf{LC\&Acc} & \textbf{Acc.}
& \textbf{LC\&Acc} & \textbf{Acc.} & \textbf{LC}
& \textbf{LC\&Acc} \\
\midrule
\multicolumn{15}{@{}l}{\textbf{DeepSeek-Distill-Qwen-1.5B}} \\
Base
& 4.5 & 18.3 & 7.3 & 17.2 & 0.1 & 23.3 & 0.1 & 13.9 & 0.4 & 9.0 & 2.5 & 16.3 & 23.0 & 32.4 \\
SFT
& 12.4 & 12.9 & 10.3 & 10.4 & 1.7 & 1.7 & 2.6 & 2.6 & 2.6 & 2.6 & 5.9 & 6.0 & 97.4 & 28.8 \\
Naive RL
& 0.0 & 40.7 & 0.0 & \textbf{38.3} & 0.0 & \textbf{34.7} & 0.0 & \textbf{28.9} & 0.0 & \textbf{28.3} & 0.0 & \textbf{34.2} & 0 & 38.5 \\
SoftLC-RL
& 21.0 & 22.9 & 21.6 & 21.6 & 14.4 & 14.4 & 0.0 & 25.9 & 10.5 & 10.5 & 13.5 & 19.1 & 78.4 & 37.3 \\
M-Thinker
& 26.0 & 26.0 & 26.5 & 26.5 & \textbf{31.0} & 31.0 & \textbf{28.5} & 28.5 & 18.5 & 18.5 & 26.1 & 26.1 & 99.8 & 32.5 \\
Direct Comb.
& 38.5 & 38.8 & 32.4 & 32.4 & 23.1 & 23.1 & 22.7 & 22.7 & 22.1 & 22.1 & 27.7 & 27.8 & 99.7 & 38.1 \\
\method{}
& \textbf{44.3} & \textbf{44.3} & \textbf{38.0} & 38.1 & 25.1 & 25.1 & 25.7 & 25.7 & \textbf{26.8} & 26.8 & \textbf{32.0} & 32.0 & \textbf{99.9} & \textbf{45.3} \\
\midrule
\multicolumn{15}{@{}l}{\textbf{Qwen3-1.7B}} \\
Base
& 0.0 & 54.2 & 0.0 & 50.9 & 0.0 & 52.8 & 0.0 & 50.7 & 0.0 & 45.5 & 0.0 & 50.8 & 0.0 & 50.1 \\
SFT
& 33.9 & 34.4 & 35.7 & 35.7 & 16.2 & 16.2 & 17.4 & 17.7 & 15.3 & 15.3 & 23.7 & 23.9 & 99.0 & 49.7 \\
Naive RL
& 0.0 & 60.9 & 0.0 & 61.1 & 0.0 & 57.7 & 0.0 & 56.0 & 0.0 & \textbf{56.7} & 0.0 & 58.5 & 0.0 & 63.5 \\
SoftLC-RL
& 54.3 & 54.6 & 51.1 & 51.1 & 43.5 & 43.5 & 36.0 & 36.2 & 40.2 & 40.2 & 45.0 & 45.1 & 99.8 & 52.0 \\
M-Thinker
& 55.3 & 55.3 & 51.3 & 51.3 & 44.9 & 44.9 & 39.1 & 39.1 & 39.5 & 39.5 & 46.0 & 46.0 & 99.9 & 61.0 \\
Direct Comb.
& 60.1 & 60.1 & 61.7 & 61.7 & 57.8 & 57.8 & 55.1 & 55.1 & 48.9 & 48.9 & 56.7 & 56.7 & 99.9 & 64.7 \\
\method{}
& \textbf{63.2} & \textbf{63.2} & \textbf{63.0} & \textbf{63.0} & \textbf{60.3} & \textbf{60.3} & \textbf{56.3} & \textbf{56.3} & \textbf{55.8} & 55.8 & \textbf{59.7} & \textbf{59.7} & \textbf{100.0} & \textbf{67.2} \\
\midrule
\multicolumn{15}{@{}l}{\textbf{Qwen3-4B}} \\
Base
& 0.0 & 63.0 & 0.0 & 65.8 & 0.0 & 62.5 & 0.0 & 58.8 & 0.3 & 59.8 & 0.1 & 62.0 & 0.1 & 63.5 \\
SFT
& 57.7 & 59.2 & 58.5 & 58.5 & 43.0 & 43.0 & 41.3 & 41.5 & 42.6 & 42.6 & 48.6 & 49.0 & 99.3 & 57.7 \\
Naive RL
& 0.0 & 77.0 & 0.0 & 76.1 & 0.0 & 74.7 & 0.0 & 74.3 & 0.0 & 75.2 & 0.0 & 75.5 & 0.0 & 77.3 \\
SoftLC-RL
& 68.3 & 68.3 & 70.9 & 71.1 & 67.6 & 67.6 & 59.8 & 59.8 & 66.3 & 66.3 & 66.6 & 66.6 & 99.8 & 55.1 \\
M-Thinker
& 70.4 & 70.4 & 71.0 & 71.0 & 65.7 & 65.7 & 65.4 & 65.4 & 64.9 & 64.9 & 67.5 & 67.5 & 99.9 & 55.5 \\
Direct Comb.
& 78.1 & 78.1 & 77.6 & 77.6 & 75.8 & 75.8 & 72.1 & 72.1 & 72.3 & 72.3 & 75.2 & 75.2 & 99.9 & 77.1 \\
\method{}
& \textbf{79.5} & \textbf{79.5} & \textbf{79.8} & \textbf{79.8} & \textbf{76.5} & \textbf{76.5} & \textbf{75.7} & \textbf{75.7} & \textbf{76.7} & \textbf{76.7} & \textbf{77.6} & \textbf{77.6} & \textbf{100.0} & \textbf{79.9} \\
\bottomrule
\end{tabular}
}
\caption{Results on MMLU-ProX STEM (\%; avg@4). The best result within each backbone and column is shown in bold.}
\label{tab:mmlu_prox_stem}
\end{table*}

\paragraph{\method{} consistently improves multilingual reasoning across backbones.}
As shown in Table~\ref{tab:main_result}, \method{} achieves the highest average LC\&Acc on all three backbones, with scores of 36.1\%, 47.4\%, and 63.7\%, respectively. It outperforms SoftLC-RL by 10.3 percentage points on average, with consistent gains across all five target languages. These results demonstrate the effectiveness of \method{} across different model scales and levels of multilingual capability.

Although Naive RL achieves relatively high raw accuracy, its language consistency drops to 1.3\% or below, indicating that it largely falls back to English reasoning. In contrast, \method{} maintains 99.9--100.0\% language consistency while achieving comparable or higher accuracy: it trails Naive RL by only 0.7 points on Qwen3-1.7B and surpasses it by 0.7 points on Qwen3-4B. Moreover, \method{} achieves the best English LC\&Acc on all three backbones, outperforming Naive RL by 3.7 points on average. Thus, its multilingual improvements neither rely on switching to English nor compromise English reasoning ability.

\paragraph{\method{} outperforms alignment-based training and direct combination.}
Notably, M-Thinker brings marginal improvements over SoftLC-RL on the Qwen3 backbones. Our analysis shows that these models already exhibit high cross-lingual reasoning-trace alignment before training, making trace-level rewards less discriminative. In contrast, \method{} improves reinforcement signals at both the question-translation and target-language reasoning stages, and therefore remains effective on models with stronger multilingual capabilities. An analysis of cross-lingual reasoning-trace alignment is provided in Appendix~\ref{app:m-thinker}.

\method{} also consistently outperforms Direct Combination, with an average gain of 4.2 percentage points across the three backbones and a gain of up to 6.0 points on DeepSeek-Distill-Qwen-1.5B. These results show that simply combining translation training with target-language RLVR is insufficient to optimize both capabilities effectively, and that translation reward refinement and reinforcement signal recovery are critical to the final performance.

\paragraph{\method{} generalizes beyond mathematical reasoning.}
Although \method{} is trained exclusively on mathematical problems, it consistently and substantially improves performance across all three backbones on the five MMLU-ProX STEM domains. As shown in Table~\ref{tab:mmlu_prox_stem}, \method{} outperforms SoftLC-RL by an average of 14.7 percentage points in LC\&Acc. These results demonstrate that the improvements learned by \method{} are not confined to the mathematical training domain, but transfer to broader multilingual STEM problems requiring diverse domain knowledge.

\section{Analysis}
\subsection{Ablation Study}
To examine the contribution of each component in \method{}, we conduct ablations on Qwen3-1.7B and report average LC\&Acc on MMATH. All variants start from the same checkpoint and use the same source questions and optimization configuration. Direct Combination retains translation training and target-language RLVR but disables translation reward refinement and reinforcement signal recovery. We then enable each mechanism individually and compare them with the full \method{}. In addition, we construct a w/o Translation Training variant that retains translation generation and verification, as well as subsequent target-language reasoning training, but excludes translation samples from policy updates. Results are shown in Table~\ref{tab:ablation}.

\begin{table}[t]
\centering
\small
\setlength{\tabcolsep}{4pt}
\renewcommand{\arraystretch}{1.1}
\begin{tabular}{lccc}
\toprule
\textbf{Method}
& \textbf{Refinement}
& \textbf{Recovery}
& \textbf{LC\&Acc} \\
\midrule
Direct Combination
& -- & -- & 44.1 \\
+ Reward Refinement
& $\checkmark$ & -- & 45.8 \\
+ Signal Recovery
& -- & $\checkmark$ & 46.5 \\
w/o Translation Training
& $\checkmark$ & $\checkmark$ & 44.6 \\
\method{}
& $\checkmark$ & $\checkmark$ & \textbf{47.4} \\
\bottomrule
\end{tabular}
\caption{\textbf{Ablation study on Qwen3-1.7B.} We report average LC\&Acc on MMATH. The w/o Translation Training variant retains refined translation rewards for data construction but excludes translation samples from policy optimization.}
\label{tab:ablation}
\end{table}

Both mechanisms improve performance when enabled individually. Reward refinement increases LC\&Acc from 44.1\% to 45.8\%, a gain of 1.7 percentage points, while signal recovery raises it to 46.5\%, a gain of 2.4 points. Combining both mechanisms yields the best performance of 47.4\%, outperforming Direct Combination by 3.3 points and further improving over either single-component variant. These results show that reward refinement and signal recovery both contribute to the final performance of \method{} and provide complementary benefits. Moreover, removing translation training reduces LC\&Acc from 47.4\% to 44.6\%, a drop of 2.8 points. This result demonstrates that directly optimizing question translation is important for improving multilingual question understanding and that the gains of \method{} do not arise solely from target-language reasoning training.

\subsection{Effectiveness of Translation Reward Refinement}

Downstream reasoning accuracy can identify task-critical translation errors that general-purpose translation metrics may overlook. However, a zero reward is ambiguous: it may indicate that the translation omits or alters crucial information, or simply that the model cannot solve the underlying problem. Without filtering, this ambiguity introduces substantial noise into translation rewards, causing semantically faithful translations to be penalized merely because the model fails to solve the problem.

English-solvability filtering reduces this interference by retaining only source problems that the model can solve reliably in English. To directly assess Reward Refinement, we use Qwen3-1.7B and add English-solvability filtering to Direct Combination, yielding the + Reward Refinement variant while keeping the remaining setup fixed. Using independent translation-quality judgments, we define a false negative as a correct translation receiving zero reward and a false positive as an incorrect translation receiving positive reward. Filtering reduces the false-negative rate from 31.3\% to 5.2\% and the false-positive rate from 5.3\% to 2.2\%, substantially removing incorrect zero rewards while keeping positive-reward noise low. Appendices~\ref{app:translation_reward_noise} and~\ref{app:threshold_analysis} provide the evaluation setup and results across different solvability thresholds.

Filtering also improves translation quality. On all MATH500 examples across the five target languages, DeepSeek-V3.2-Exp prefers translations generated by + Reward Refinement over those generated by Direct Combination in 56.3\% of cases, while preferring Direct Combination in 15.6\%. We aggregate multiple judgments with swapped candidate orders to reduce position bias; the remaining 28.1\% are ties. On the out-of-domain FLORES-200 benchmark, Reward Refinement increases COMET from 82.1 to 84.7. These controlled comparisons show that a more reliable translation reward improves the preservation of task-critical information in mathematical problems and also benefits general-domain translation. The improvement carries over to downstream reasoning: as shown in Table~\ref{tab:ablation}, Reward Refinement increases average LC\&Acc on MMATH from 44.1\% to 45.8\%. Thus, English-solvability filtering not only reduces reward noise but also makes translation training more useful for downstream multilingual reasoning.

\subsection{Effectiveness of Target-Language Signal Recovery}

When all $G$ directly sampled target-language responses to a problem are incorrect, they receive the same reward, leaving GRPO without a discriminative signal for learning correct reasoning. To examine how often this occurs and whether Signal Recovery remains useful throughout optimization, we analyze the training dynamics of Qwen3-1.7B and Qwen3-4B. We track the proportion of target-language problems for which all $G$ direct responses are incorrect, which we call the all-failure rate. Among these problems, recovery is considered successful if hint-assisted sampling produces at least one correct response.

\begin{figure}[t]
    \centering
    \includegraphics[width=\linewidth]{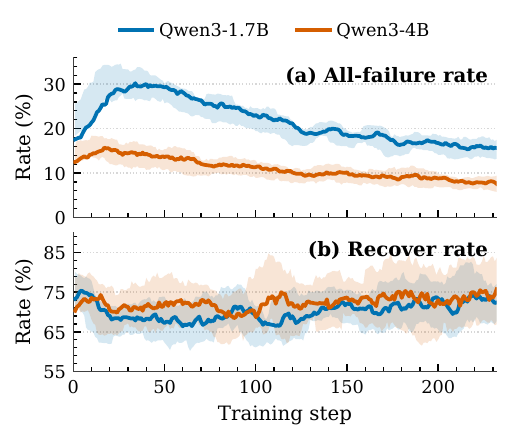}
    \caption{
    \textbf{Training dynamics of target-language signal recovery.}
    (a) Proportion of target-language problems with no correct response among $G$ direct samples.
    (b) Proportion of these problems recovered by hint-assisted sampling.
    }
    \label{fig:signal_recovery_dynamics}
\end{figure}

As shown in Figure~\ref{fig:signal_recovery_dynamics}, the all-failure rate initially increases and then declines on both models. This pattern is consistent with the interaction between translation and target-language reasoning. Early in training, improved translation allows more target-language problems to pass translation verification and enter target-language RLVR. Some of these problems remain solvable but difficult for the model to solve directly in the target language, increasing the all-failure rate. As the model's target-language reasoning ability improves, it can solve more verified problems directly, and the rate declines. Thus, better question understanding initially broadens the set of usable target-language problems, while target-language reasoning gradually adapts to this expanded training set. Appendix~\ref{app:stage1_retention} reports the corresponding expansion of the Stage-1 training pool.

Despite the changing all-failure rate, the recovery success rate remains approximately 70--80\% for both models. Thus, \method{} finds at least one correct target-language response for most problems on which direct sampling fails, converting uninformative groups into usable RLVR data. Similar dynamics on Qwen3-1.7B and Qwen3-4B show that Signal Recovery remains effective across training stages and model scales.

Recovered signals also improve downstream performance: adding Signal Recovery to Direct Combination raises average MMATH LC\&Acc from 44.1\% to 46.5\% on Qwen3-1.7B. Since evaluation uses no reasoning hints, this gain shows that hint-assisted training transfers to unassisted target-language inference.

\section{Conclusion}
We present \method{}, which improves target-language question understanding and reasoning with capability-specific reinforcement signals. English-solvability filtering refines reasoning-based translation rewards, while self-generated English hints recover target-language RLVR signals when direct sampling fails. Across three backbones, \method{} improves language-consistent accuracy on MMATH and MMLU-ProX while maintaining near-perfect language consistency; ablations confirm that both mechanisms contribute.

\section*{Limitations}
Although \method{} consistently improves performance across three backbones and five target languages, the scope of our evaluation can be further expanded. First, we do not evaluate extremely low-resource languages; future work will examine a broader range of language-resource levels and language families. Second, the largest model evaluated in this work has 4B parameters, and the effectiveness of \method{} on larger reasoning models remains to be investigated.

%
%

\bibliography{main}

@article{comet,
  title={COMET: A Neural Framework for MT Evaluation},
  author={Ricardo Rei and Craig Alan Stewart and Ana C. Farinha and Alon Lavie},
  journal={ArXiv},
  year={2020},
  volume={abs/2009.09025},
  url={https://api.semanticscholar.org/CorpusID:221819581}
}

@misc{lang,
      title={LANG: Reinforcement Learning for Multilingual Reasoning with Language-Adaptive Hint Guidance}, 
      author={Yuchun Fan and Bei Li and Peiguang Li and Yilin Wang and Yongyu Mu and Jian Yang and Xin Chen and Rongxiang Weng and Jingang Wang and Xunliang Cai and Jingbo Zhu and Tong Xiao},
      year={2026},
      eprint={2605.22567},
      archivePrefix={arXiv},
      primaryClass={cs.CL},
      url={https://arxiv.org/abs/2605.22567}, 
}

@misc{tapo,
      title={TAPO: Translation Augmented Policy Optimization for Multilingual Mathematical Reasoning}, 
      author={Xu Huang and Zhejian Lai and Zixian Huang and Jiajun Chen and Shujian Huang},
      year={2026},
      eprint={2603.25419},
      archivePrefix={arXiv},
      primaryClass={cs.CL},
      url={https://arxiv.org/abs/2603.25419}, 
}

@misc{mmath,
      title={MMATH: A Multilingual Benchmark for Mathematical Reasoning}, 
      author={Wenyang Luo and Wayne Xin Zhao and Jing Sha and Shijin Wang and Ji-Rong Wen},
      year={2025},
      eprint={2505.19126},
      archivePrefix={arXiv},
      primaryClass={cs.CL},
      url={https://arxiv.org/abs/2505.19126}, 
}

@misc{mistralai2025magistral,
      title={Magistral}, 
      author={Mistral-AI and : and Abhinav Rastogi and Albert Q. Jiang and Andy Lo and Gabrielle Berrada and Guillaume Lample and Jason Rute and Joep Barmentlo and Karmesh Yadav and Kartik Khandelwal and Khyathi Raghavi Chandu and Léonard Blier and Lucile Saulnier and Matthieu Dinot and Maxime Darrin and Neha Gupta and Roman Soletskyi and Sagar Vaze and Teven Le Scao and Yihan Wang and Adam Yang and Alexander H. Liu and Alexandre Sablayrolles and Amélie Héliou and Amélie Martin and Andy Ehrenberg and Anmol Agarwal and Antoine Roux and Arthur Darcet and Arthur Mensch and Baptiste Bout and Baptiste Rozière and Baudouin De Monicault and Chris Bamford and Christian Wallenwein and Christophe Renaudin and Clémence Lanfranchi and Darius Dabert and Devon Mizelle and Diego de las Casas and Elliot Chane-Sane and Emilien Fugier and Emma Bou Hanna and Gauthier Delerce and Gauthier Guinet and Georgii Novikov and Guillaume Martin and Himanshu Jaju and Jan Ludziejewski and Jean-Hadrien Chabran and Jean-Malo Delignon and Joachim Studnia and Jonas Amar and Josselin Somerville Roberts and Julien Denize and Karan Saxena and Kush Jain and Lingxiao Zhao and Louis Martin and Luyu Gao and Lélio Renard Lavaud and Marie Pellat and Mathilde Guillaumin and Mathis Felardos and Maximilian Augustin and Mickaël Seznec and Nikhil Raghuraman and Olivier Duchenne and Patricia Wang and Patrick von Platen and Patryk Saffer and Paul Jacob and Paul Wambergue and Paula Kurylowicz and Pavankumar Reddy Muddireddy and Philomène Chagniot and Pierre Stock and Pravesh Agrawal and Romain Sauvestre and Rémi Delacourt and Sanchit Gandhi and Sandeep Subramanian and Shashwat Dalal and Siddharth Gandhi and Soham Ghosh and Srijan Mishra and Sumukh Aithal and Szymon Antoniak and Thibault Schueller and Thibaut Lavril and Thomas Robert and Thomas Wang and Timothée Lacroix and Valeriia Nemychnikova and Victor Paltz and Virgile Richard and Wen-Ding Li and William Marshall and Xuanyu Zhang and Yunhao Tang},
      year={2025},
      eprint={2506.10910},
      archivePrefix={arXiv},
      primaryClass={cs.CL},
      url={https://arxiv.org/abs/2506.10910}, 
}

@misc{park2025crosslingualcollapselanguagecentricfoundation,
      title={Cross-lingual Collapse: How Language-Centric Foundation Models Shape Reasoning in Large Language Models}, 
      author={Cheonbok Park and Jeonghoon Kim and Joosung Lee and Sanghwan Bae and Jaegul Choo and Kang Min Yoo},
      year={2025},
      eprint={2506.05850},
      archivePrefix={arXiv},
      primaryClass={cs.CL},
      url={https://arxiv.org/abs/2506.05850}, 
}

@misc{grpo,
      title={DeepSeekMath: Pushing the Limits of Mathematical Reasoning in Open Language Models}, 
      author={Zhihong Shao and Peiyi Wang and Qihao Zhu and Runxin Xu and Junxiao Song and Xiao Bi and Haowei Zhang and Mingchuan Zhang and Y. K. Li and Y. Wu and Daya Guo},
      year={2024},
      eprint={2402.03300},
      archivePrefix={arXiv},
      primaryClass={cs.CL},
      url={https://arxiv.org/abs/2402.03300}, 
}

@misc{polymath,
      title={PolyMath: Evaluating Mathematical Reasoning in Multilingual Contexts}, 
      author={Yiming Wang and Pei Zhang and Jialong Tang and Haoran Wei and Baosong Yang and Rui Wang and Chenshu Sun and Feitong Sun and Jiran Zhang and Junxuan Wu and Qiqian Cang and Yichang Zhang and Fei Huang and Junyang Lin and Fei Huang and Jingren Zhou},
      year={2025},
      eprint={2504.18428},
      archivePrefix={arXiv},
      primaryClass={cs.CL},
      url={https://arxiv.org/abs/2504.18428}, 
}

@misc{mmluprox,
      title={MMLU-ProX: A Multilingual Benchmark for Advanced Large Language Model Evaluation}, 
      author={Weihao Xuan and Rui Yang and Heli Qi and Qingcheng Zeng and Yunze Xiao and Aosong Feng and Dairui Liu and Yun Xing and Junjue Wang and Fan Gao and Jinghui Lu and Yuang Jiang and Huitao Li and Xin Li and Kunyu Yu and Ruihai Dong and Shangding Gu and Yuekang Li and Xiaofei Xie and Felix Juefei-Xu and Foutse Khomh and Osamu Yoshie and Qingyu Chen and Douglas Teodoro and Nan Liu and Randy Goebel and Lei Ma and Edison Marrese-Taylor and Shijian Lu and Yusuke Iwasawa and Yutaka Matsuo and Irene Li},
      year={2025},
      eprint={2503.10497},
      archivePrefix={arXiv},
      primaryClass={cs.CL},
      url={https://arxiv.org/abs/2503.10497}, 
}

@misc{qwen3,
      title={Qwen3 Technical Report}, 
      author={An Yang and Anfeng Li and Baosong Yang and Beichen Zhang and Binyuan Hui and Bo Zheng and Bowen Yu and Chang Gao and Chengen Huang and Chenxu Lv and Chujie Zheng and Dayiheng Liu and Fan Zhou and Fei Huang and Feng Hu and Hao Ge and Haoran Wei and Huan Lin and Jialong Tang and Jian Yang and Jianhong Tu and Jianwei Zhang and Jianxin Yang and Jiaxi Yang and Jing Zhou and Jingren Zhou and Junyang Lin and Kai Dang and Keqin Bao and Kexin Yang and Le Yu and Lianghao Deng and Mei Li and Mingfeng Xue and Mingze Li and Pei Zhang and Peng Wang and Qin Zhu and Rui Men and Ruize Gao and Shixuan Liu and Shuang Luo and Tianhao Li and Tianyi Tang and Wenbiao Yin and Xingzhang Ren and Xinyu Wang and Xinyu Zhang and Xuancheng Ren and Yang Fan and Yang Su and Yichang Zhang and Yinger Zhang and Yu Wan and Yuqiong Liu and Zekun Wang and Zeyu Cui and Zhenru Zhang and Zhipeng Zhou and Zihan Qiu},
      year={2025},
      eprint={2505.09388},
      archivePrefix={arXiv},
      primaryClass={cs.CL},
      url={https://arxiv.org/abs/2505.09388}, 
}

@misc{thinkNatively,
      title={Think Natively: Unlocking Multilingual Reasoning with Consistency-Enhanced Reinforcement Learning}, 
      author={Xue Zhang and Yunlong Liang and Fandong Meng and Songming Zhang and Kaiyu Huang and Yufeng Chen and Jinan Xu and Jie Zhou},
      year={2025},
      eprint={2510.07300},
      archivePrefix={arXiv},
      primaryClass={cs.CL},
      url={https://arxiv.org/abs/2510.07300}, 
}

@misc{mapo,
      title={MAPO: Advancing Multilingual Reasoning through Multilingual Alignment-as-Preference Optimization}, 
      author={Shuaijie She and Wei Zou and Shujian Huang and Wenhao Zhu and Xiang Liu and Xiang Geng and Jiajun Chen},
      year={2024},
      eprint={2401.06838},
      archivePrefix={arXiv},
      primaryClass={cs.CL},
      url={https://arxiv.org/abs/2401.06838}, 
}

@misc{breakingLanguageBarrier,
      title={Breaking Language Barriers in Multilingual Mathematical Reasoning: Insights and Observations}, 
      author={Nuo Chen and Zinan Zheng and Ning Wu and Ming Gong and Dongmei Zhang and Jia Li},
      year={2024},
      eprint={2310.20246},
      archivePrefix={arXiv},
      primaryClass={cs.CL},
      url={https://arxiv.org/abs/2310.20246}, 
}

@misc{zhu2024questiontranslationtrainingbetter,
      title={Question Translation Training for Better Multilingual Reasoning}, 
      author={Wenhao Zhu and Shujian Huang and Fei Yuan and Shuaijie She and Jiajun Chen and Alexandra Birch},
      year={2024},
      eprint={2401.07817},
      archivePrefix={arXiv},
      primaryClass={cs.CL},
      url={https://arxiv.org/abs/2401.07817}, 
}

@misc{deepseekr1,
      title={DeepSeek-R1: Incentivizing Reasoning Capability in LLMs via Reinforcement Learning}, 
      author={{DeepSeek-AI} and Daya Guo and Dejian Yang and Haowei Zhang and Junxiao Song and Ruoyu Zhang and Runxin Xu and Qihao Zhu and Shirong Ma and Peiyi Wang and Xiao Bi and Xiaokang Zhang and Xingkai Yu and Yu Wu and Z. F. Wu and Zhibin Gou and Zhihong Shao and Zhuoshu Li and Ziyi Gao and Aixin Liu and Bing Xue and Bingxuan Wang and Bochao Wu and Bei Feng and Chengda Lu and Chenggang Zhao and Chengqi Deng and Chenyu Zhang and Chong Ruan and Damai Dai and Deli Chen and Dongjie Ji and Erhang Li and Fangyun Lin and Fucong Dai and Fuli Luo and Guangbo Hao and Guanting Chen and Guowei Li and H. Zhang and Han Bao and Hanwei Xu and Haocheng Wang and Honghui Ding and Huajian Xin and Huazuo Gao and Hui Qu and Hui Li and Jianzhong Guo and Jiashi Li and Jiawei Wang and Jingchang Chen and Jingyang Yuan and Junjie Qiu and Junlong Li and J. L. Cai and Jiaqi Ni and Jian Liang and Jin Chen and Kai Dong and Kai Hu and Kaige Gao and Kang Guan and Kexin Huang and Kuai Yu and Lean Wang and Lecong Zhang and Liang Zhao and Litong Wang and Liyue Zhang and Lei Xu and Leyi Xia and Mingchuan Zhang and Minghua Zhang and Minghui Tang and Meng Li and Miaojun Wang and Mingming Li and Ning Tian and Panpan Huang and Peng Zhang and Qiancheng Wang and Qinyu Chen and Qiushi Du and Ruiqi Ge and Ruisong Zhang and Ruizhe Pan and Runji Wang and R. J. Chen and R. L. Jin and Ruyi Chen and Shanghao Lu and Shangyan Zhou and Shanhuang Chen and Shengfeng Ye and Shiyu Wang and Shuiping Yu and Shunfeng Zhou and Shuting Pan and S. S. Li and Shuang Zhou and Shaoqing Wu and Shengfeng Ye and Tao Yun and Tian Pei and Tianyu Sun and T. Wang and Wangding Zeng and Wanjia Zhao and Wen Liu and Wenfeng Liang and Wenjun Gao and Wenqin Yu and Wentao Zhang and W. L. Xiao and Wei An and Xiaodong Liu and Xiaohan Wang and Xiaokang Chen and Xiaotao Nie and Xin Cheng and Xin Liu and Xin Xie and Xingchao Liu and Xinyu Yang and Xinyuan Li and Xuecheng Su and Xuheng Lin and X. Q. Li and Xiangyue Jin and Xiaojin Shen and Xiaosha Chen and Xiaowen Sun and Xiaoxiang Wang and Xinnan Song and Xinyi Zhou and Xianzu Wang and Xinxia Shan and Y. K. Li and Y. Q. Wang and Y. X. Wei and Yang Zhang and Yanhong Xu and Yao Li and Yao Zhao and Yaofeng Sun and Yaohui Wang and Yi Yu and Yichao Zhang and Yifan Shi and Yiliang Xiong and Ying He and Yishi Piao and Yisong Wang and Yixuan Tan and Yiyang Ma and Yiyuan Liu and Yongqiang Guo and Yuan Ou and Yuduan Wang and Yue Gong and Yuheng Zou and Yujia He and Yunfan Xiong and Yuxiang Luo and Yuxiang You and Yuxuan Liu and Yuyang Zhou and Y. X. Zhu and Yanhong Xu and Yanping Huang and Yaohui Li and Yi Zheng and Yuchen Zhu and Yunxian Ma and Ying Tang and Yukun Zha and Yuting Yan and Z. Z. Ren and Zehui Ren and Zhangli Sha and Zhe Fu and Zhean Xu and Zhenda Xie and Zhengyan Zhang and Zhewen Hao and Zhicheng Ma and Zhigang Yan and Zhiyu Wu and Zihui Gu and Zijia Zhu and Zijun Liu and Zilin Li and Ziwei Xie and Ziyang Song and Zizheng Pan and Zhen Huang and Zhipeng Xu and Zhongyu Zhang and Zhen Zhang},
      year={2025},
      eprint={2501.12948},
      archivePrefix={arXiv},
      primaryClass={cs.CL},
      url={https://arxiv.org/abs/2501.12948}, 
}

@misc{o3,
      title={Competitive Programming with Large Reasoning Models}, 
      author={OpenAI and : and Ahmed El-Kishky and Alexander Wei and Andre Saraiva and Borys Minaiev and Daniel Selsam and David Dohan and Francis Song and Hunter Lightman and Ignasi Clavera and Jakub Pachocki and Jerry Tworek and Lorenz Kuhn and Lukasz Kaiser and Mark Chen and Max Schwarzer and Mostafa Rohaninejad and Nat McAleese and o3 contributors and Oleg Mürk and Rhythm Garg and Rui Shu and Szymon Sidor and Vineet Kosaraju and Wenda Zhou},
      year={2025},
      eprint={2502.06807},
      archivePrefix={arXiv},
      primaryClass={cs.LG},
      url={https://arxiv.org/abs/2502.06807}, 
}

@misc{hwang2025learngloballyspeaklocally,
      title={Learn Globally, Speak Locally: Bridging the Gaps in Multilingual Reasoning}, 
      author={Jaedong Hwang and Kumar Tanmay and Seok-Jin Lee and Ayush Agrawal and Hamid Palangi and Kumar Ayush and Ila Fiete and Paul Pu Liang},
      year={2025},
      eprint={2507.05418},
      archivePrefix={arXiv},
      primaryClass={cs.CL},
      url={https://arxiv.org/abs/2507.05418}, 
}

@inproceedings{Qi_2025,
   title={When Models Reason in Your Language: Controlling Thinking Language Comes at the Cost of Accuracy},
   url={http://dx.doi.org/10.18653/v1/2025.findings-emnlp.1103},
   DOI={10.18653/v1/2025.findings-emnlp.1103},
   booktitle={Findings of the Association for Computational Linguistics: EMNLP 2025},
   publisher={Association for Computational Linguistics},
   author={Qi, Jirui and Chen, Shan and Xiong, Zidi and Fernández, Raquel and Bitterman, Danielle and Bisazza, Arianna},
   year={2025},
   pages={20279–20296} }

@misc{ko2025understandsolvetranslatebridging,
      title={Understand, Solve and Translate: Bridging the Multilingual Mathematical Reasoning Gap}, 
      author={Hyunwoo Ko and Guijin Son and Dasol Choi},
      year={2025},
      eprint={2501.02448},
      archivePrefix={arXiv},
      primaryClass={cs.CL},
      url={https://arxiv.org/abs/2501.02448}, 
}

@misc{dapo-math,
      title={DAPO: An Open-Source LLM Reinforcement Learning System at Scale}, 
      author={Qiying Yu and Zheng Zhang and Ruofei Zhu and Yufeng Yuan and Xiaochen Zuo and Yu Yue and Weinan Dai and Tiantian Fan and Gaohong Liu and Lingjun Liu and Xin Liu and Haibin Lin and Zhiqi Lin and Bole Ma and Guangming Sheng and Yuxuan Tong and Chi Zhang and Mofan Zhang and Wang Zhang and Hang Zhu and Jinhua Zhu and Jiaze Chen and Jiangjie Chen and Chengyi Wang and Hongli Yu and Yuxuan Song and Xiangpeng Wei and Hao Zhou and Jingjing Liu and Wei-Ying Ma and Ya-Qin Zhang and Lin Yan and Mu Qiao and Yonghui Wu and Mingxuan Wang},
      year={2025},
      eprint={2503.14476},
      archivePrefix={arXiv},
      primaryClass={cs.LG},
      url={https://arxiv.org/abs/2503.14476}, 
}

@misc{kang2026multilingualreasoninggapsemerge,
      title={Why Do Multilingual Reasoning Gaps Emerge in Reasoning Language Models?}, 
      author={Deokhyung Kang and Seonjeong Hwang and Daehui Kim and Hyounghun Kim and Gary Geunbae Lee},
      year={2026},
      eprint={2510.27269},
      archivePrefix={arXiv},
      primaryClass={cs.CL},
      url={https://arxiv.org/abs/2510.27269}, 
}

\clearpage
\appendix
\setcounter{secnumdepth}{2}
\section{Complete Training Procedure}
\label{app:algorithm}

Algorithm~\ref{alg:trit} gives the complete training procedure of \method{}.
Stage 1 first samples responses to all English source questions. These rollouts
provide English RLVR data and identify the subset of questions that the model
can solve reliably. The model then generates target-language translations only
for this subset and verifies them through English reasoning, thereby refining
the translation rewards. Stage 2 operates only on positively verified
translations, which reduces interference from question misunderstanding and
insufficient problem-solving ability in target-language RLVR. It first performs
direct target-language sampling; when every direct response is incorrect, it
uses self-generated English hints to recover an effective signal. English
reasoning, translation, and retained target-language reasoning samples are
finally used in the same GRPO update.

\begin{algorithm*}[!t]
\footnotesize
\caption{Training Procedure of \method{}}
\label{alg:trit}
\begin{algorithmic}[1]
\STATE \textbf{Input:} English questions $\mathcal{Q}_{\mathrm{en}}$, target language $L_{\mathrm{tgt}}$, filtering threshold $\theta$, rollout sizes $G$ and $K$
\FOR{each training iteration $s=1,\ldots,S$}
    \STATE Initialize $\mathcal{D}_{\mathrm{en}},\mathcal{D}_{\mathrm{trans}},\mathcal{D}_{\mathrm{tgt}},\mathcal{Q}_{\mathrm{stable}},\mathcal{Q}_{\mathrm{verified}}\leftarrow\emptyset$

    \STATE \textbf{// Stage 1: Translation RL Signal Refinement}
    \FOR{$q_{\mathrm{en}}\in\mathcal{Q}_{\mathrm{en}}$}
        \STATE Sample English responses $\{e_i\}_{i=1}^{G}\sim\pi_s(\cdot\mid q_{\mathrm{en}},L_{\mathrm{en}})$; compute $\mathrm{AvgAcc}(q_{\mathrm{en}})=\frac{1}{G}\sum_i r_{\mathrm{acc}}(e_i)$
        \STATE $\mathcal{D}_{\mathrm{en}}\leftarrow\mathcal{D}_{\mathrm{en}}\cup\{(q_{\mathrm{en}},e_i,r_{\mathrm{acc}}(e_i))\}_{i=1}^{G}$
        \STATE \textbf{If} $\mathrm{AvgAcc}(q_{\mathrm{en}})\geq\theta$: $\mathcal{Q}_{\mathrm{stable}}\leftarrow\mathcal{Q}_{\mathrm{stable}}\cup\{q_{\mathrm{en}}\}$
    \ENDFOR

    \FOR{$q_{\mathrm{en}}\in\mathcal{Q}_{\mathrm{stable}}$}
        \STATE Translate $q_{\mathrm{en}}$ by sampling $\{t_j\}_{j=1}^{K}\sim\pi_s(\cdot\mid q_{\mathrm{en}},L_{\mathrm{tgt}})$
        \FOR{$j=1,\ldots,K$}
            \IF{$\neg\operatorname{Valid}(t_j)$}
                \STATE $r_{\mathrm{trans}}^{j}\leftarrow 0$
            \ELSE
                \STATE Sample English verification responses $\{v_i^j\}_{i=1}^{G}\sim\pi_s(\cdot\mid t_j,L_{\mathrm{en}})$
                \STATE $r_{\mathrm{trans}}^{j}\leftarrow\mathbb{I}\!\left[\frac{1}{G}\sum_i r_{\mathrm{acc}}(v_i^j)>0\right]$
                \IF{$r_{\mathrm{trans}}^{j}=1$}
                    \STATE $\mathcal{H}_j\leftarrow\{\operatorname{Prefix}_{\approx 200}(v_i^j)\mid r_{\mathrm{acc}}(v_i^j)=1\}$ \COMMENT{end at nearest sentence boundary}
                    \STATE $\mathcal{Q}_{\mathrm{verified}}\leftarrow\mathcal{Q}_{\mathrm{verified}}\cup\{(t_j,\mathcal{H}_j)\}$
                \ENDIF
            \ENDIF
        \ENDFOR
        \STATE $\mathcal{D}_{\mathrm{trans}}\leftarrow\mathcal{D}_{\mathrm{trans}}\cup\{(q_{\mathrm{en}},t_j,r_{\mathrm{trans}}^{j})\}_{j=1}^{K}$
    \ENDFOR

    \STATE \textbf{// Stage 2: Target-Language RL Signal Refinement and Recovery}
    \FOR{$(t_j,\mathcal{H}_j)\in\mathcal{Q}_{\mathrm{verified}}$}
        \STATE Sample direct responses $\{o_i\}_{i=1}^{G}\sim\pi_s(\cdot\mid t_j,L_{\mathrm{tgt}})$
        \IF{$\sum_i r_{\mathrm{acc}}(o_i)>0$}
            \STATE $\mathcal{D}_{\mathrm{tgt}}\leftarrow\mathcal{D}_{\mathrm{tgt}}\cup\{(t_j,o_i,r_{\mathrm{final}}(o_i))\}_{i=1}^{G}$
        \ELSE
            \STATE Sample one hint $h\sim\operatorname{Uniform}(\mathcal{H}_j)$
            \STATE Sample hint-assisted responses $\{\widetilde{o}_i\}_{i=1}^{G}\sim\pi_s(\cdot\mid t_j,h,L_{\mathrm{tgt}})$
            \IF{$\sum_i r_{\mathrm{acc}}(\widetilde{o}_i)>0$}
                \STATE $\mathcal{D}_{\mathrm{tgt}}\leftarrow\mathcal{D}_{\mathrm{tgt}}\cup\{((t_j,h),\widetilde{o}_i,r_{\mathrm{final}}(\widetilde{o}_i))\}_{i=1}^{G}$
            \ENDIF
        \ENDIF
    \ENDFOR

    \STATE $\pi_{s+1}\leftarrow\operatorname{GRPO}\!\left(\pi_s;\mathcal{D}_{\mathrm{en}}\cup\mathcal{D}_{\mathrm{trans}}\cup\mathcal{D}_{\mathrm{tgt}}\right)$
\ENDFOR
\end{algorithmic}
\end{algorithm*}

\section{Implementation Details}
\label{app:training_details}

\subsection{Training Data}
\label{app:data_construction}

We use DAPO-MATH-17K~\cite{dapo-math} as the source of English training
problems and construct training instances for French, Portuguese, Japanese,
Korean, and Thai. The RL set contains 3,000 English source problems per target
language. To cover a broad range of difficulty, 2,000 problems have an initial
English reasoning accuracy in $(0, 0.5)$, while the remaining 1,000 have zero
initial accuracy.
\method{} itself generates target-language translations and English reasoning
hints online from these English problems.

Some baselines require pre-constructed multilingual data. For these methods,
we translate the English source questions with DeepSeek-V3.2-Exp and use
Qwen3-32B to verify translation fidelity through multiple evaluation passes.
For the SFT baseline, Qwen3-32B additionally generates target-language
responses, from which we retain 3,000 correct and language-consistent examples
per language. These externally generated translations and responses are used
only by the corresponding baselines and are not used to train \method{}.

\newpage
\subsection{Cold Start for DeepSeek-Distill-Qwen-1.5B}

DeepSeek-Distill-Qwen-1.5B has substantially weaker initial target-language
generation ability than the Qwen3 backbones. Before RL training, we therefore
perform a cold-start stage that teaches this backbone the target-language
reasoning pattern. Qwen3-8B is prompted to answer English questions directly in
the target language, without translating the input question, and we retain
responses that are both correct and language-consistent. The resulting
cold-start checkpoint is shared by \method{} and all RL baselines using this
backbone, so the comparison among RL methods starts from the same model. No
cold-start stage is used for Qwen3-1.7B or Qwen3-4B.

\subsection{Training Configuration}

We optimize all models with AdamW. For the DeepSeek cold-start stage, we use a
batch size of 64, a learning rate of $1\times10^{-5}$, two epochs, linear
warmup, and cosine learning-rate decay. For RL, we use a global batch size of
512, a mini-batch size of 64, a learning rate of $1\times10^{-6}$, and a KL
coefficient of $0.001$. Unless otherwise stated, we sample $G=6$ responses for
each reasoning task and $K=4$ candidate translations. The maximum sequence
length is 8,192 tokens. English reasoning hints target approximately 200 tokens
and end at the complete-sentence boundary closest to that length.

\subsection{Translation Evaluation Protocol}
\label{app:judge_protocols}

For analyses that require translation-quality judgments, we use
DeepSeek-V3.2-Exp and aggregate multiple independent judgments by majority
vote. The judge receives the English source question, the candidate
translation, and, when available, the benchmark-provided target-language
reference. For pairwise comparisons, we reverse the candidate order across
evaluation passes to reduce position bias. The COMET agreement analysis uses
GPT-5 as an independent judge, as specified in its corresponding subsection.
These protocols are used only for analysis; they do not provide feedback
during \method{} training.

\section{Detailed Results on Mathematical Reasoning}
\label{app:difficulty}

MMATH combines mathematical problems with substantially different difficulty
levels. To complement the aggregate results in the main paper, we report
detailed results for the high-difficulty AIME24/25 problems and the
medium-difficulty MATH500 problems on Qwen3-1.7B. In addition to Accuracy,
Language Consistency, and LC\&Acc, we report degenerate repetition rates to
characterize a common failure of target-language reasoning.

\begin{table*}[t]
\centering
{\small
\setlength{\tabcolsep}{1mm}
\begin{tabular}{@{}l*{5}{cc}cccc@{}}
\toprule
& \multicolumn{2}{c}{\textbf{FR}}
& \multicolumn{2}{c}{\textbf{PT}}
& \multicolumn{2}{c}{\textbf{JA}}
& \multicolumn{2}{c}{\textbf{KO}}
& \multicolumn{2}{c}{\textbf{TH}}
& \multicolumn{4}{c}{\textbf{Non-EN Avg.}} \\
\cmidrule(lr){2-3}\cmidrule(lr){4-5}\cmidrule(lr){6-7}
\cmidrule(lr){8-9}\cmidrule(lr){10-11}\cmidrule(lr){12-15}
\textbf{Method}
& \textbf{LC\&Acc} & \textbf{Acc.}
& \textbf{LC\&Acc} & \textbf{Acc.}
& \textbf{LC\&Acc} & \textbf{Acc.}
& \textbf{LC\&Acc} & \textbf{Acc.}
& \textbf{LC\&Acc} & \textbf{Acc.}
& \textbf{LC\&Acc} & \textbf{Acc.} & \textbf{LC} & \textbf{Rep.} \\
\midrule
\multicolumn{15}{@{}l}{\textbf{Qwen3-1.7B}} \\
Base
& 0.0 & 23.8 & 0.0 & 25.0 & 0.0 & 20.6 & 0.0 & 23.9 & 0.0 & 21.1
& 0.0 & 22.9 & 0.0 & 4.2 \\
SFT
& 20.0 & 20.0 & 19.4 & 19.4 & 10.5 & 10.5 & 8.9 & 9.4 & 11.6 & 11.6
& 14.1 & 14.2 & 98.2 & 3.4 \\
Naive RL
& 0.0 & 31.1 & 0.0 & 30.0 & 0.0 & \textbf{28.9} & 0.0 & \textbf{25.0} & 0.0 & \textbf{27.8}
& 0.0 & \textbf{28.6} & 0.0 & 3.8 \\
SoftLC-RL
& 23.3 & 23.3 & 23.3 & 23.3 & 12.2 & 12.2 & 15.0 & 15.0 & 18.3 & 18.3
& 18.4 & 18.4 & \textbf{100.0} & 30.7 \\
M-Thinker
& 21.1 & 21.1 & 23.9 & 23.9 & 12.2 & 12.2 & 11.7 & 11.7 & 15.6 & 15.6
& 16.9 & 16.9 & 99.8 & 44.4 \\
Direct Comb.
& 28.9 & 28.9 & 33.3 & 33.3 & 19.4 & 19.4 & \textbf{18.9} & 18.9 & 21.1 & 21.1
& 24.3 & 24.3 & 99.9 & 12.0 \\
\method{}
& \textbf{37.8} & \textbf{37.8} & \textbf{37.8} & \textbf{37.8} & \textbf{28.3} & 28.3 & 15.0 & 15.0 & \textbf{21.7} & 21.7
& \textbf{28.1} & 28.1 & \textbf{100.0} & 9.8 \\
\bottomrule
\end{tabular}
}
\caption{Detailed results on AIME24/25 with Qwen3-1.7B (\%; avg@4). The best reported Acc. and LC\&Acc in each column are shown in bold. Rep. denotes the repetition rate.}
\label{tab:aime_results}
\end{table*}

\begin{table*}[t]
\centering
{\small
\setlength{\tabcolsep}{1mm}
\begin{tabular}{@{}l*{5}{cc}cccc@{}}
\toprule
& \multicolumn{2}{c}{\textbf{FR}}
& \multicolumn{2}{c}{\textbf{PT}}
& \multicolumn{2}{c}{\textbf{JA}}
& \multicolumn{2}{c}{\textbf{KO}}
& \multicolumn{2}{c}{\textbf{TH}}
& \multicolumn{4}{c}{\textbf{Non-EN Avg.}} \\
\cmidrule(lr){2-3}\cmidrule(lr){4-5}\cmidrule(lr){6-7}
\cmidrule(lr){8-9}\cmidrule(lr){10-11}\cmidrule(lr){12-15}
\textbf{Method}
& \textbf{LC\&Acc} & \textbf{Acc.}
& \textbf{LC\&Acc} & \textbf{Acc.}
& \textbf{LC\&Acc} & \textbf{Acc.}
& \textbf{LC\&Acc} & \textbf{Acc.}
& \textbf{LC\&Acc} & \textbf{Acc.}
& \textbf{LC\&Acc} & \textbf{Acc.} & \textbf{LC} & \textbf{Rep.} \\
\midrule
\multicolumn{15}{@{}l}{\textbf{Qwen3-1.7B}} \\
Base
& 0.0 & 86.1 & 0.0 & 86.9 & 0.0 & 83.9 & 0.0 & 82.5 & 0.0 & 82.2
& 0.0 & 84.3 & 0.0 & 0.9 \\
SFT
& 81.0 & 82.7 & 81.6 & 82.0 & 65.9 & 66.0 & 61.0 & 61.2 & 63.0 & 63.3
& 70.5 & 71.0 & 99.1 & 21.5 \\
Naive RL
& 0.0 & 90.7 & 0.0 & 89.6 & 0.0 & \textbf{86.0} & 0.0 & \textbf{88.3} & 0.0 & \textbf{87.8}
& 0.0 & \textbf{88.5} & 0.0 & 1.0 \\
SoftLC-RL
& 87.3 & 87.3 & 88.1 & 88.1 & 77.0 & 77.0 & 74.4 & 74.4 & 77.3 & 77.7
& 80.8 & 80.9 & 99.9 & 6.9 \\
M-Thinker
& 87.1 & 87.5 & 86.9 & 87.4 & 75.6 & 75.8 & 72.4 & 72.4 & 77.8 & 78.1
& 80.0 & 80.2 & 99.6 & 10.0 \\
Direct Comb.
& 91.0 & 91.1 & 90.8 & 90.8 & 83.9 & 83.9 & 83.1 & 83.1 & 84.9 & 84.9
& 86.7 & 86.8 & \textbf{100.0} & 2.6 \\
\method{}
& \textbf{92.0} & \textbf{92.0} & \textbf{91.2} & \textbf{91.2} & \textbf{84.8} & 84.8 & \textbf{83.6} & 83.6 & \textbf{86.3} & 86.3
& \textbf{87.6} & 87.6 & 99.9 & 1.6 \\
\bottomrule
\end{tabular}
}
\caption{Detailed results on MATH500 with Qwen3-1.7B (\%; avg@4). The best reported Acc. and LC\&Acc in each column are shown in bold. Rep. denotes the repetition rate.}
\label{tab:math500_results}
\end{table*}

\paragraph{Results on difficult problems.}
On AIME24/25, \method{} obtains 28.1\% average LC\&Acc while maintaining
100.0\% language consistency, outperforming SoftLC-RL and M-Thinker by 9.7 and
11.2 percentage points, respectively. It also improves over Direct Combination
by 3.8 points.

\paragraph{Results on medium-difficulty problems.}
On MATH500, \method{} reaches 87.6\% average LC\&Acc with 99.9\% language
consistency, outperforming SoftLC-RL and M-Thinker by 6.8 and 7.6 percentage
points, respectively.

\paragraph{Generation stability.}
Compared with Direct Combination, \method{} reduces the repetition rate from
12.0\% to 9.8\% on AIME24/25. The same pattern holds on MATH500, where the
rate decreases from 2.6\% to 1.6\%.

\section{Further Analysis of Translation Reward Refinement}
\label{app:noise_analysis}

\subsection{Why General-Purpose Translation Metrics Are Insufficient}

Mathematical questions are sensitive to small semantic changes in terminology,
quantities, and logical constraints. A translation can therefore remain fluent
and lexically similar to a reference while changing the problem that must be
solved. The main paper gives a representative example in which COMET assigns a
higher score to a Korean translation that weakens
``parallelogram'' to ``quadrilateral'' than to the precise translation, even
though the missing geometric constraint changes the problem.

We quantify this limitation by comparing COMET preferences with judgments from
GPT-5 on paired translations. We sample 200 pairs from FLORES-200 general text
and another 200 pairs from DAPO-MATH mathematical problems. COMET agrees with
the independent judge on 59\% of the general-domain pairs but only 46\% of the
mathematical pairs. The decrease suggests that a general-purpose translation
metric is less reliable when translation quality depends on preserving
task-critical mathematical semantics.

\subsection{Reasoning Accuracy as a Translation-Quality Signal}
\label{app:reason-accuracy}

We next examine whether downstream reasoning outcomes provide a more
task-sensitive signal. For each MATH500 source question, we generate multiple
candidate translations and compare pairs that lead to different English
reasoning accuracies. DeepSeek-V3.2-Exp judges which candidate more faithfully
preserves the source question using the protocol in
Appendix~\ref{app:judge_protocols}.

\begin{figure}[t]
    \centering
    \includegraphics[width=\linewidth]{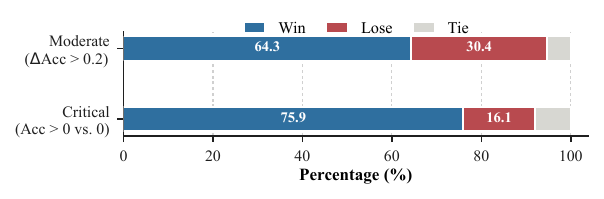}
    \caption{
        Relationship between translation quality and reasoning accuracy.
        Win, lose, and tie are defined from the perspective of the candidate
        with higher downstream English reasoning accuracy. Moderate denotes
        pairs with an accuracy difference greater than 0.2; Critical denotes
        pairs in which one candidate yields zero accuracy and the other yields
        nonzero accuracy.
    }
    \label{fig:translation_reasoning}
\end{figure}

As shown in Figure~\ref{fig:translation_reasoning}, when the reasoning-accuracy
difference exceeds 0.2, the higher-accuracy translation is preferred in
64.3\% of pairs, whereas it loses in 30.4\%. The relationship becomes stronger
for critical pairs: when one translation produces zero accuracy and the other
produces nonzero accuracy, the latter is preferred in 75.9\% of cases and loses
in only 16.1\%. These results do not imply that reasoning accuracy is a perfect
translation metric, but they show that it captures whether a translation
preserves information needed to solve the problem. This motivates its use as a
task-oriented translation reward.

\subsection{Binary versus Graded Translation Rewards}
\label{app:graded_translation_reward}

The preceding analysis raises a natural question: whether the magnitude of
reasoning accuracy can provide a more informative reward than the binary
signal used by \method{}. We therefore evaluate a graded variant that compares
the verification accuracy of a translation with the Stage-1 English accuracy
of its source problem. Let
$a_j=\operatorname{AvgAcc}(t_j)$ and
$a_{\mathrm{en}}=\operatorname{AvgAcc}(q_{\mathrm{en}})$. The graded reward is
defined as

\begin{equation}
r_{\mathrm{graded}}^{j}=
\begin{cases}
1, & a_j>a_{\mathrm{en}}-0.2,\\
0.5, & 0<a_j\leq a_{\mathrm{en}}-0.2,\\
0, & a_j=0.
\end{cases}
\end{equation}

This reward distinguishes translations that retain accuracy close to that of
the English source from those that remain solvable but incur a larger accuracy
drop. Table~\ref{tab:reward_granularity} compares it with the binary reward
under the same Qwen3-1.7B setting.

\begin{table}[H]
\centering
{\small
\setlength{\tabcolsep}{5pt}
\begin{tabular}{lc}
\toprule
Translation Reward & Non-English Avg. Acc. \\
\midrule
Graded reward & 46.8 \\
Binary reward (\method{}) & \textbf{47.4} \\
\bottomrule
\end{tabular}
}
\caption{
Comparison of translation-reward granularity on MMATH with
Qwen3-1.7B.}
\label{tab:reward_granularity}
\end{table}

The graded variant obtains 46.8\% average accuracy on MMATH, compared with
47.4\% for the binary reward. One plausible explanation is that both
accuracies are estimated from only $G$ sampled responses. Their exact
magnitudes can therefore be noisy, and comparing two empirical estimates
introduces additional variance. The binary reward instead asks only whether a
translation preserves enough information to produce at least one correct
reasoning trajectory, making it more robust under a limited verification
budget.

\subsection{Translation Reward Noise}
\label{app:translation_reward_noise}

The binary translation reward is computed from the model's English reasoning
outcomes on a translated problem and can therefore err in both directions. We
call a reward a \emph{false negative} when an independently verified correct
translation receives zero reward, and a \emph{false positive} when an
independently verified incorrect translation receives positive reward.

We compare Direct Combination, which applies no English-solvability filtering,
with Reward Refinement under several filtering thresholds on Qwen3-1.7B. All
settings use the same source data and evaluation protocol; Reward Refinement
retains only English source problems whose sampled English accuracy reaches
$\theta$. Translation correctness is determined using the protocol in
Appendix~\ref{app:judge_protocols}.

\begin{table}[t]
\centering
{\small
\setlength{\tabcolsep}{5pt}
\begin{tabular}{lcc}
\toprule
\textbf{Setting} & \textbf{FN (\%)} & \textbf{FP (\%)} \\
\midrule
No filtering & 31.3 & 5.3 \\
$\theta=1/6$ & 9.3 & 3.1 \\
$\theta=1/3$ (default) & 5.2 & 2.2 \\
$\theta=1/2$ & \textbf{3.8} & \textbf{2.0} \\
\bottomrule
\end{tabular}
}
\caption{
Noise in the binary translation reward on Qwen3-1.7B.
A false negative (FN) is an independently verified correct translation that
receives zero reward; a false positive (FP) is an independently verified
incorrect translation that receives positive reward. Lower values are better.
}
\label{tab:translation_reward_noise}
\end{table}

As shown in Table~\ref{tab:translation_reward_noise}, English-solvability
filtering consistently reduces reward noise as the threshold increases. The
false-negative rate decreases from 31.3\% without filtering to 9.3\% at
$\theta=1/6$, 5.2\% at the default $\theta=1/3$, and 3.8\% at $\theta=1/2$.
Although the filtering mechanism is designed primarily to resolve ambiguous
zero rewards, the false-positive rate also decreases from 5.3\% without
filtering to 3.1\%, 2.2\%, and 2.0\%, respectively. False positives are much
less frequent than false negatives in all settings. This may partly reflect our
difficulty-focused training pool: for challenging problems, a semantically
incorrect translation is unlikely to yield the correct answer by chance. At the
default threshold, filtering reduces the false-negative and false-positive
rates by 26.1 and 3.1 percentage points, respectively. The primary benefit is
therefore the removal of incorrect zero rewards, while positive-reward noise
remains low. A stricter threshold further improves reward reliability, although
reward noise alone does not determine the best threshold because stronger
filtering also reduces training coverage.

\subsection{Sensitivity to the English-Solvability Threshold}
\label{app:threshold_analysis}

The threshold $\theta$ controls the trade-off between reward reliability and
training-data coverage. We use $\theta=1/3$ as the default and conduct a
post-hoc sensitivity analysis to examine this trade-off. When $\theta=0$, all English source problems enter
translation training, including problems that the model cannot currently
solve. Raising the threshold removes many such problems and makes zero
translation rewards less ambiguous. If the threshold becomes too strict,
however, too few problems remain for effective translation training.

\begin{figure}[t]
    \centering
    \includegraphics[width=\linewidth]{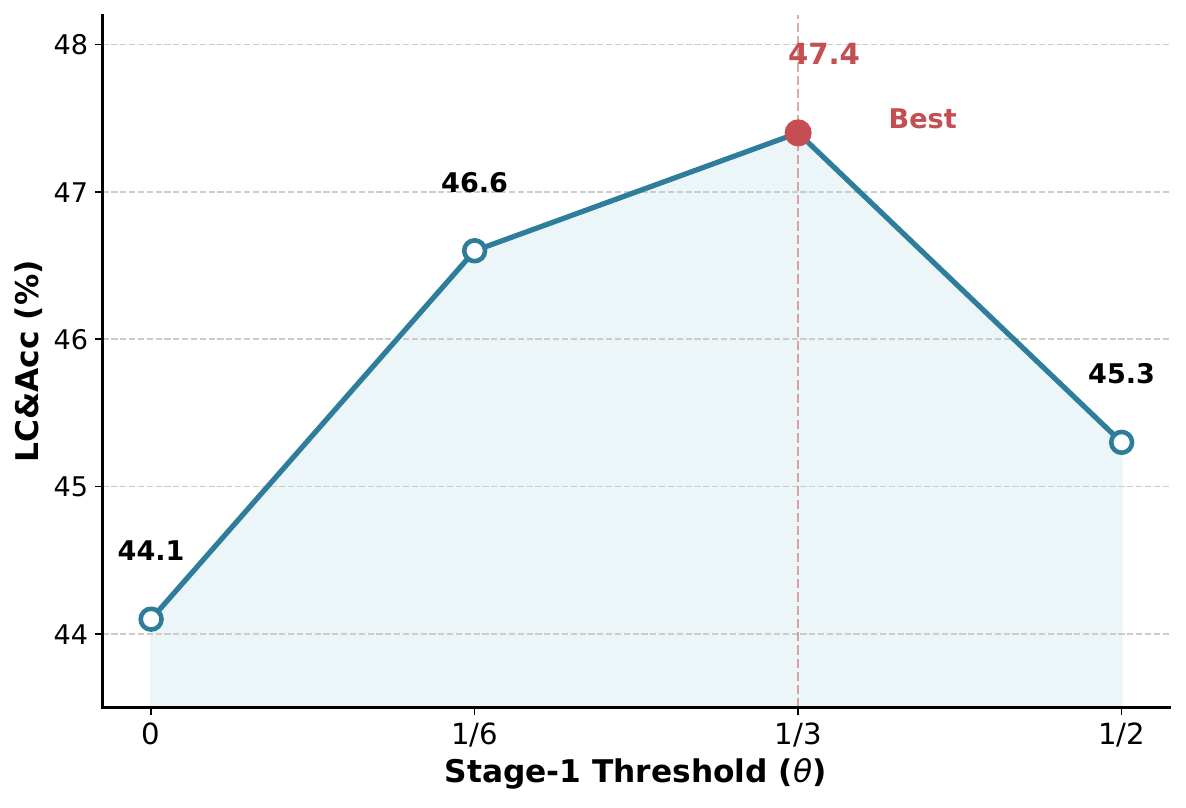}
    \caption{
        Sensitivity to the English-solvability threshold.
        Average LC\&Acc on MMATH with Qwen3-1.7B under different values of
        $\theta$. A moderate threshold improves reward reliability without
        excessively reducing translation-training coverage.
    }
    \label{fig:stage1_threshold}
\end{figure}

Figure~\ref{fig:stage1_threshold} shows this pattern on Qwen3-1.7B. Performance
increases from 44.1\% LC\&Acc without filtering to 46.6\% at $\theta=1/6$ and
reaches its maximum of 47.4\% at $\theta=1/3$. Increasing the threshold to
$1/2$ further reduces reward noise but lowers LC\&Acc to 45.3\%, indicating
that the loss of training coverage outweighs the additional improvement in
reward reliability. These results support the use of $\theta=1/3$, which
provides a favorable balance between reward reliability and training-data
coverage.

\subsection{Expansion of the Stage-1 Training Pool}
\label{app:stage1_retention}

To examine whether English-solvability filtering confines translation training
to a fixed subset of initially solvable problems, we track the Stage-1 pass
ratio during training. We define this ratio as the proportion of English source
problems in each batch that satisfy $\operatorname{AvgAcc}(q_{\mathrm{en}})\geq
\theta$ and therefore enter translation training.

\begin{figure}[t]
    \centering
    \includegraphics[width=\linewidth]{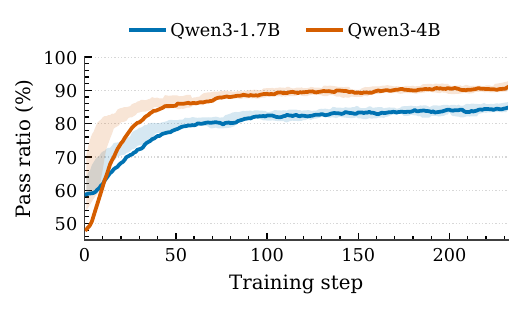}
    \caption{
        Stage-1 English-solvability pass ratio during training.
        Both models progressively expand the pool of source problems eligible
        for translation training.
    }
    \label{fig:stage1_filtering}
\end{figure}

As shown in Figure~\ref{fig:stage1_filtering}, the eligible training pool
expands substantially for both models. Qwen3-1.7B starts at approximately 59\%
and gradually stabilizes near 85\%, whereas Qwen3-4B starts lower at about 48\%
but increases more rapidly, overtakes Qwen3-1.7B at around step 15, and
stabilizes near 90--91\% after 100 steps. This pattern is consistent with the
role of English RLVR in improving source-problem solvability: as training
progresses, previously filtered problems become eligible for translation
training. English-solvability filtering therefore maintains reward reliability
without permanently restricting training to the model's initially solvable
problems.

\section{Additional Analyses}

\subsection{Hint Length}
\label{app:hint_length}

We evaluate nominal hint lengths of approximately 100, 200, and 400 tokens. In
each setting, the prefix ends at the complete-sentence boundary closest to the
target length rather than at an exact token count. As shown in
Table~\ref{tab:hint_length}, LC\&Acc varies by only 0.5 percentage points across
the three settings, indicating that signal recovery is robust to hint length
within this range. We use approximately 200 tokens by default to balance the
amount of guidance and input length.

\begin{table}[H]
\centering
{\small
\setlength{\tabcolsep}{5pt}
\begin{tabular}{lc}
\toprule
Nominal Hint Length (tokens) & Non-English Avg. LC\&Acc \\
\midrule
Approx. 100 & 46.9 \\
Approx. 200 (default) & 47.4 \\
Approx. 400 & 47.1 \\
\bottomrule
\end{tabular}
}
\caption{
Sensitivity to hint length on MMATH with Qwen3-1.7B.
}
\label{tab:hint_length}
\end{table}

\subsection{Why M-Thinker Provides Limited Gains on Qwen3}
\label{app:m-thinker}

M-Thinker improves target-language reasoning by explicitly rewarding alignment
between target-language and English reasoning traces. On the Qwen3 backbones,
however, it provides only marginal improvements over SoftLC-RL. We examine this
behavior using cross-lingual thinking alignment (CTA), measured on MMATH with
DeepSeek-V3.2-Exp as the trace-alignment judge.

\begin{table}[htbp]
\centering
{\small
\setlength{\tabcolsep}{5pt}
\renewcommand{\arraystretch}{1.12}
\begin{tabular}{lc}
\toprule
\textbf{Method} & \textbf{CTA Score} \\
\midrule
Baseline (Qwen3-1.7B) & 0.930 \\
\midrule
M-Thinker & 0.923 \\
\method (Ours) & \textbf{0.947} \\
\bottomrule
\end{tabular}
}
\caption{
    Cross-lingual thinking alignment (CTA) analysis.
    CTA scores are measured on MMATH using DeepSeek-V3.2-Exp to judge alignment with English traces.
    Higher scores indicate stronger alignment.
}
\label{tab:cta_analysis}
\end{table}

As shown in Table~\ref{tab:cta_analysis}, Qwen3-1.7B already has a CTA score of
0.930 before training. M-Thinker changes this score only slightly and in fact
reduces it to 0.923. When trace alignment is already high, a trace-level
alignment reward offers limited discrimination among sampled responses.
\method{} instead improves capability-specific reinforcement signals at the
question-translation and target-language reasoning stages, and raises CTA to
0.947 without directly optimizing this metric.

\FloatBarrier


\section{Reward Modeling and Implementation}
\label{app:repetition}

For English reasoning and translation verification, answer correctness is
determined by the benchmark's verifiable answer. For target-language reasoning,
we use a reward function with four components:

\begin{itemize}
    \item \textbf{Accuracy reward ($\text{r}_{\text{acc}}$):}
    $\text{r}_{\text{acc}}=1$ if the answer is correct, otherwise 0.

    \item \textbf{Language consistency reward ($\text{r}_{\text{lang}}$):}
    We use \textit{langdetect}\footnote{\url{https://github.com/Mimino666/langdetect}}
    to check both the reasoning trace and final answer.
    $\text{r}_{\text{lang}}=1$ only if both are in the target language,
    otherwise 0.

    \item \textbf{Repetition penalty ($\text{r}_{\text{rep}}$):}
    $\text{r}_{\text{rep}}=1$ if no degenerate repetition is detected,
    otherwise 0.

    \item \textbf{Format reward ($\text{r}_{\text{fmt}}$):}
    $\text{r}_{\text{fmt}}=1$ if the output follows the required format,
    otherwise 0.
\end{itemize}

We adopt a compositional reward structure in which correctness is rewarded only
when all quality constraints are satisfied. Specifically,

\[
\begin{aligned}
\text{r}_{\text{final}} =
\begin{cases}
1,   & \text{if } C \land (\text{r}_{\text{acc}} = 1),\\[4pt]
0.1, & \text{if } C \land (\text{r}_{\text{acc}} = 0),\\[4pt]
0,   & \text{otherwise},
\end{cases}
\end{aligned}
\]
\[
C = (\text{r}_{\text{fmt}} = 1 \land \text{r}_{\text{lang}} = 1 \land \text{r}_{\text{rep}} = 1).
\]

This design rewards correct answers only when the output is well-formed,
language-consistent, and free of repetition, while assigning a small reward to
well-formed, language-consistent, non-repetitive responses with incorrect
answers.

\subsection{Repetition Detection}

We detect repetition with two complementary rules. First, a response is marked
as repetitive if it contains a repeated 20-gram. Second, it is marked as
repetitive if a line containing at least 20 tokens occurs six or more times.
The first rule captures local phrase-level loops, while the second captures
longer repeated reasoning blocks. These checks are applied only to the sampled
response, not to the English hint included in the input.

\section{Translation Prompt}

\begin{figure}[t]
    \centering
    \begin{tcolorbox}[
        enhanced,
        colback=gray!5!white,      
        frame hidden,              
        borderline west={3pt}{0pt}{gray!70!black}, 
        sharp corners,             
        top=10pt, bottom=10pt, left=10pt, right=10pt,
        drop shadow={black!30!white,opacity=0.2},
        breakable
    ]
        \small
        You are a professional translator.
        
        \vspace{2mm}
        \textbf{Task:} Translate the given English question into \texttt{\{language\}} with zero loss of mathematical or logical meaning.
        
        \vspace{2mm}
        \textbf{Output format (MUST keep tags):} \\
        \texttt{<TRANSLATION>} \textit{\{language\} text only} \texttt{</TRANSLATION>}
        
        \vspace{2mm}
        \textbf{Rules:} \\
        \textbf{1.} Re-state any numbers, symbols, units exactly as they appear. \\
        \textbf{2.} Keep LaTeX (\texttt{\$...\$}) unchanged; do not translate inside \texttt{\$...\$}. \\
        \textbf{3.} If the question contains multiple-choice items (A), (B), ... keep the same labels. \\
        \textbf{4.} Use natural \texttt{\{language\}} wording, but stay one-to-one faithful to the original semantics. \\
        \textbf{5.} Reply with ONLY the tagged translation---no explanations.
        
        \vspace{2mm}
        \texttt{<ENGLISH QUESTION>} \textit{\{question\}} \texttt{</ENGLISH QUESTION>}
        
        \vspace{2mm}
        Please translate the above English question into \texttt{\{language\}}.
    \end{tcolorbox}
    
    \caption{
        Translation prompt template.
        The prompt asks the model to generate semantically faithful translations while preserving mathematical notation and output formatting.
    }
    \label{fig:translation_prompt}
\end{figure}

\end{document}